\pdfoutput=1

\documentclass[11pt]{article}

\usepackage{ACL2023}

\usepackage{times}
\usepackage{latexsym}

\usepackage[T1]{fontenc}

\usepackage{amsmath,amsfonts,bm}

\def\eqref#1{equation~\ref{#1}}

\def\1{\bm{1}}

\DeclareMathAlphabet{\mathsfit}{\encodingdefault}{\sfdefault}{m}{sl}
\SetMathAlphabet{\mathsfit}{bold}{\encodingdefault}{\sfdefault}{bx}{n}

\usepackage{hyperref}
\usepackage{url}
\usepackage{enumitem} 
\usepackage{graphicx}
\usepackage{multicol}
\usepackage{multirow}
\usepackage{booktabs}
\usepackage{tcolorbox}
\usepackage{amsmath}
\usepackage{wrapfig}
\usepackage{mathtools}
\usepackage[utf8]{inputenc}
\usepackage{minitoc}
\newif\ifshowcomment
\showcommenttrue 

\newcommand{\vpara}[1]{\vspace{0.05in}\noindent \textbf{#1 }}

\newcommand{\benchmark}{\texttt{DataSciBench}}
\title{DataSciBench: An LLM Agent Benchmark for Data Science}

\author{Dan Zhang$^{1,2,\dagger}$,\ Sining Zhoubian$^{1,2,\dagger}$,\ Min Cai$^{2,\dagger}$,\ Fengzu Li$^{1}$, \ Lekang Yang$^{1}$\\
\textbf{Wei Wang$^{1}$, Tianjiao Dong$^{3, \dagger}$,\ Ziniu Hu$^{4}$,\ Jie Tang$^{1}$, Yisong Yue$^{4}$}\\
$^1$Tsinghua University; $^2$Zhipu AI;\\
$^3$University of California, Berkeley; $^4$California Institute of Technology\\
\normalsize\rule{0pt}{1em}\url{https://datascibench.github.io/}\\
}
\begin{document}

\maketitle

\begin{abstract}
This paper presents \benchmark, a comprehensive benchmark for evaluating Large Language Model (LLM) capabilities in data science.
Recent related benchmarks have primarily focused on single tasks, easily obtainable ground truth, and straightforward evaluation metrics, which limits the scope of tasks that can be evaluated.
In contrast, \benchmark$\space$ is constructed based on a more comprehensive and curated collection of natural and challenging prompts for uncertain ground truth and evaluation metrics. 
We develop a semi-automated pipeline for generating ground truth (GT) and validating evaluation metrics. This pipeline utilizes and implements an LLM-based self-consistency and human verification strategy to produce accurate GT by leveraging collected prompts, predefined task types, and aggregate functions (metrics). Furthermore, we propose an innovative Task - Function - Code (TFC) framework to assess each code execution outcome based on precisely defined metrics and programmatic rules.
Our experimental framework involves testing \textbf{6} API-based models, \textbf{8} open-source general models, and \textbf{9} open-source code generation models using the diverse set of prompts we have gathered.
This approach aims to provide a more comprehensive and rigorous evaluation of LLMs in data science, revealing their strengths and weaknesses.
Experimental results demonstrate that API-based models outperform open-sourced models on all metrics and Deepseek-Coder-33B-Instruct achieves the highest score among open-sourced models.
We release all code and data at~\url{https://github.com/THUDM/DataSciBench/}.

\end{abstract}

{\let\thefootnote\relax\footnotetext{$^\dagger$ work done while these authors interned at Zhipu AI.}}

\section{Introduction}
\label{sec: introduction}
\begin{figure}[t!]
    \centering
    \includegraphics[width=0.93\linewidth]{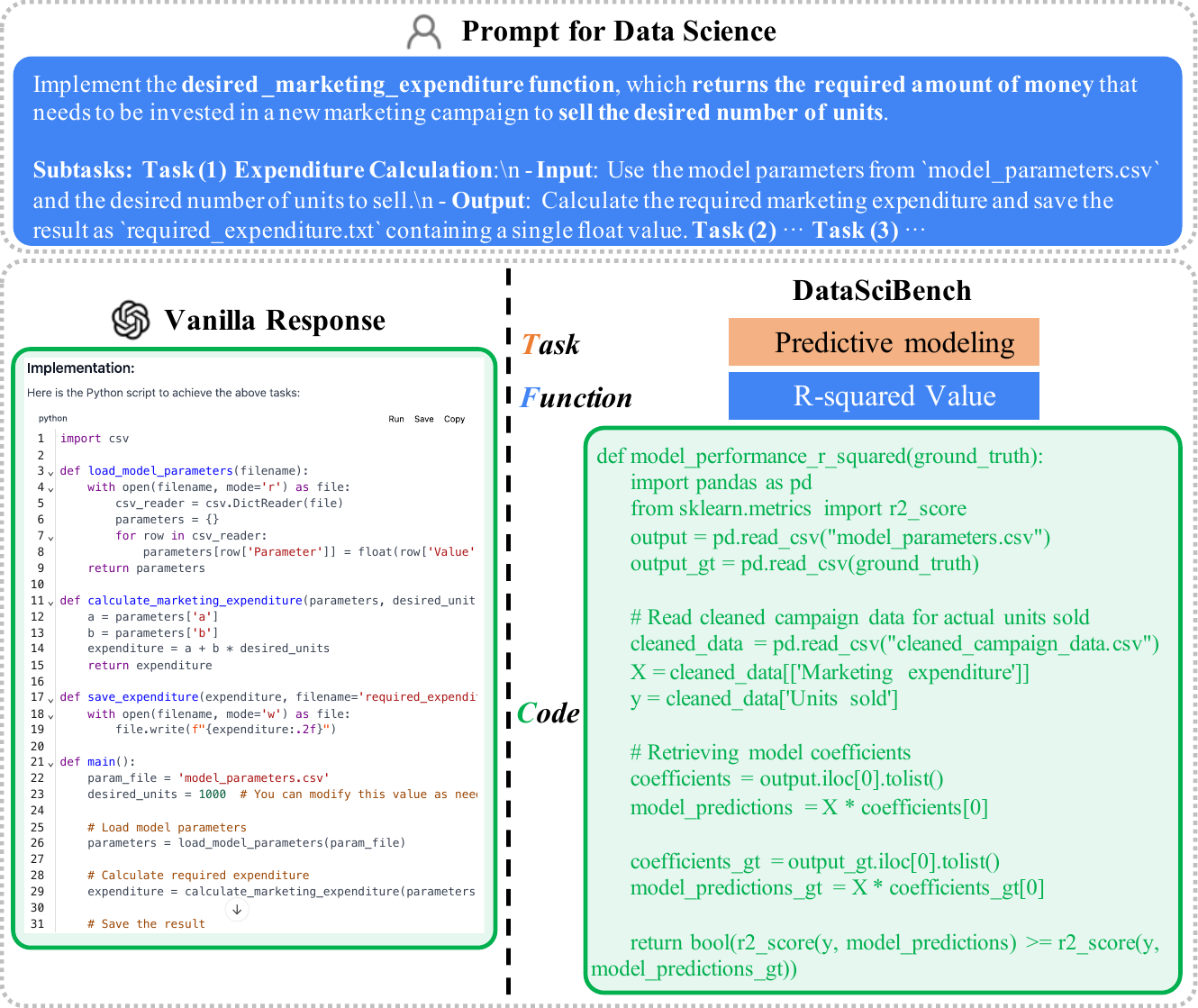}
    \vspace{-0.2cm}
    \caption{This example compares the vanilla response and the \benchmark\space response for a given prompt. The vanilla response provides only code, lacking evaluation metrics. In contrast, \benchmark\space identifies evaluation tasks, provides evaluation functions, and generates programmatic code to form a TFC list.}
    \label{fig: motivation}
\end{figure}

Large language models (LLMs)~\citep{achiam2023gpt, team2023gemini, glm2024chatglm} are increasingly used in data science and scientific domains, e.g., data analysis~\citep{hong2024datainterpreter}, protein generation~\citep{jumper2021highly, chen2024xtrimopglm}, and scientific discovery~\citep{lu2024ai} and reasoning~\citep{zhang2024sciglm, zhang2024rest}.
For data science tasks, given a publicly known problem, LLMs offer the potential to (semi-)autonomously conduct data analysis~\citep{huang2023mlbenchmarking} and data visualization~\citep{hong2024datainterpreter} by invoking code interpreters with corresponding Python libraries.
These works are benchmarked on relatively straightforward tasks where ground truth (GT) labels can be precisely obtained.
However, much of real-world data analysis requires reasoning over more complex scenarios~\citep{chen2024viseval} as shown in Figure~\ref{fig: motivation}, such as calculating expenditure, and evaluating the quality of the images generated by the data visualization task.
Properly evaluating these more complex data science tasks remains an open research direction.

\begin{table*}[t!]
    \centering
    \caption{Comparison with related work. LC denotes LeetCode.}
    \vspace{-0.3cm}
    \label{tab: comparison}
    \resizebox{0.97\textwidth}{!}{
    \begin{tabular}{c|c|c}
        \specialrule{.16em}{0pt}{.65ex}
        Benchmark & Prompt Source & Evaluation Metrics \\
        \specialrule{.10em}{.4ex}{.65ex}
        DS-1000~\citep{lai2023ds} & StackOverflow & Test Cases +
Surface-Form Constraints\\
        MLAgentBench~\citep{huang2023mlbenchmarking} & Kaggle & Acc. + Success Rate, Human \\
        LiveCodeBench~\citep{jain2024livecodebench} & LC \& AtCoder \& CodeForces & Test Cases + Pass Rate \\
        NaturalCodeBench~\citep{zhang2024naturalcodebench} & CodeGeeX & Test Cases + Pass Rate\\
        BigCodeBench~\citep{zhuo2024bigcodebench} & StackOverflow &  Test Cases + Pass Rate \\
        \specialrule{.10em}{.4ex}{.65ex} 
        Text2Analysis~\citep{he2024text2analysis} & Human \& LLM & Executable code ratio, Acc., Regression scores \\
        InfiAgent-DABench~\citep{huinfiagent} & LLM & Acc. \\
        \specialrule{.10em}{.4ex}{.65ex} 
        \benchmark$\space$ (Ours) & CodeGeeX \& BCB \& Human  & Aggregate Metrics and Programmatic Rules \\
        \specialrule{.16em}{.4ex}{0pt}
    \end{tabular}
    }
\end{table*}

While some existing benchmarks are used to evaluate LLMs for related challenges (see Table~\ref{tab: comparison}), those benchmarks (e.g., MLAgentBench, Text2Analysis) typically focus on evaluating narrower tasks with easy-to-obtain ground truth and straightforward evaluation metrics (e.g., Acc.).
For example, MLAgentBench~\citep{huang2023mlbenchmarking} presents a machine learning research benchmark by building an LLM Agent pipeline. 
As shown in Figure~\ref{fig: motivation}, the frontier of LLM evaluation is towards more complex real-world tasks that consist of multiple subtasks. For these challenging prompts, how to generate GT and define specific evaluation metrics for each subtask in a comprehensive perspective is a question worth exploring.

To address the gap between task selection, evaluation function definition, and automated code execution in data science contexts, we introduce a new benchmark, called \benchmark, which evaluates the data science abilities of LLMs and provides insights to help LLMs improve their data analysis and data visualization abilities.
Regarding collected prompts, their corresponding responses, and evaluation metrics, we hope that prompts meet the following characteristics:
1) Require more natural, challenging, and high-quality prompts to promote the development of LLMs' improvement;
2) Identify \textbf{6} defined data science tasks, e.g., data preprocessing, statistics, visualization, mining, and interpretability;
3) Require multiple types of results to perform comprehensive evaluations and distinguish models well.

In this paper, to efficiently generate GT and the evaluation metric for key tasks, we propose a novel semi-automated framework called \underline{\textbf{T}}ask - \underline{\textbf{F}}unction - \underline{\textbf{C}}ode (TFC) to address critical challenges in the automated evaluation of data science tasks.
Specifically, from a coarse-grained perspective, we first aggregate the scope of task types, evaluation functions, and corresponding codes.
Then, from a fine-grained perspective, we define programmatic rules for each function's output based on the specific tasks and compare generation results with ground truth to ensure a fair and consistent assessment. 

To validate the effectiveness of LLMs on collected \textbf{222} comprehensive prompts with proposed \textbf{519} GTs, we experiment with \textbf{6} API-based models, \textbf{8} open-sourced general models, and \textbf{9} open-sourced code generation models.
We observe that API-based models greatly outperform open-sourced models on average.
Specifically, GPT-4o surpasses all other models on all metrics and Deepseek-Coder-33B-Instruct achieves the highest score among open-sourced models. 
However, all models have significant room for improvement in following fine-grained instructions, calling the appropriate tools, executing accurate plans, and exporting the required execution outputs.

\begin{figure*}[t!]
    \centering
    \includegraphics[width=0.96\linewidth]{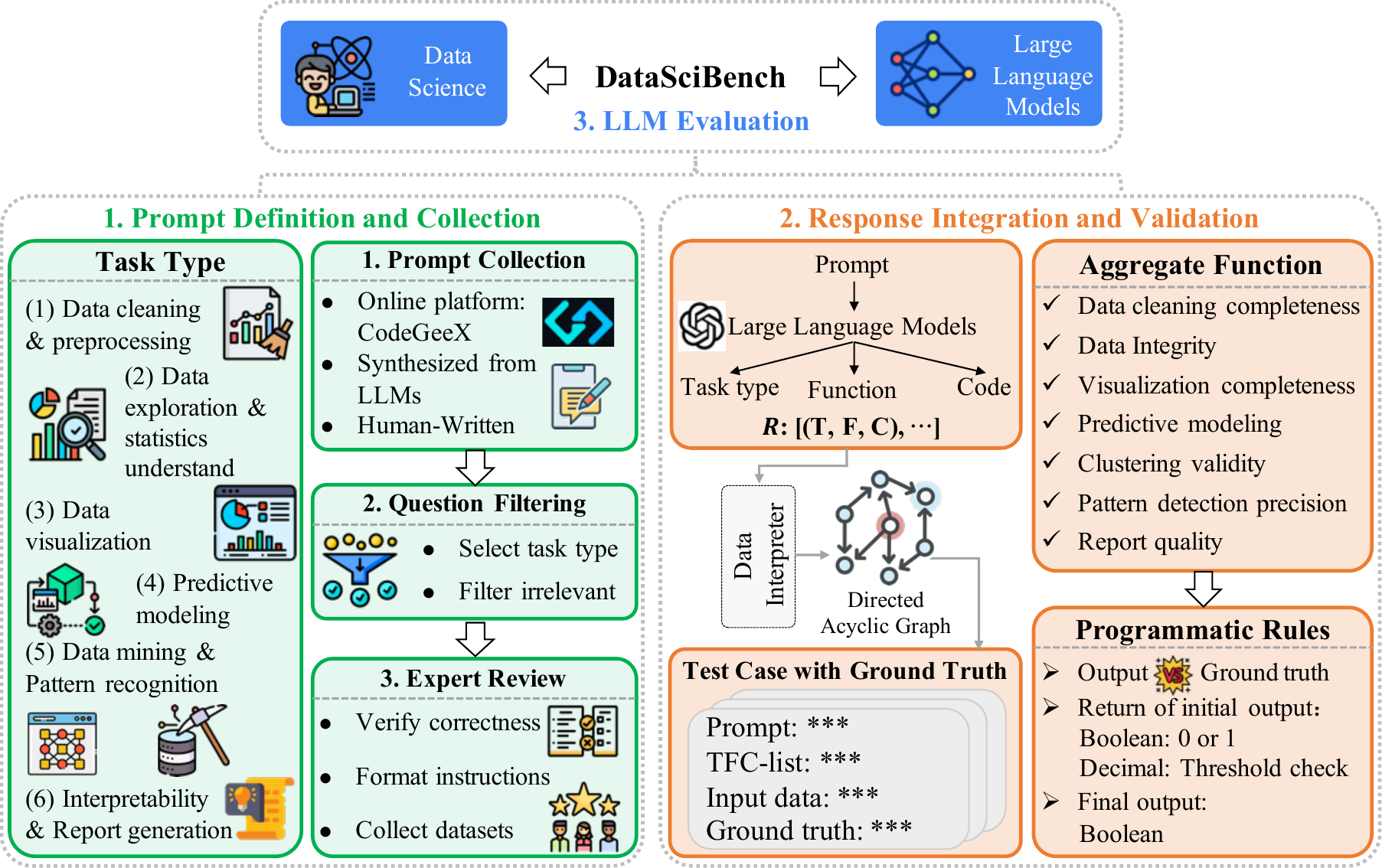}
    \vspace{-0.2cm}
    \caption{ 
    The overall framework of \benchmark$\space$ encompasses three key components aligned with Section~\ref{sec: DataSciBench}:
    1) Prompt definition and collection, which covers \textbf{6} task types, prompt collection, question filtering, and expert review.
    2) Response integration and validation, incorporating the TFC framework (\textbf{25} aggregated functions and programmatic rules) and \textbf{519} test cases with ground truth.
    3) LLM evaluation involving \textbf{23} LLMs.
    }
    \label{fig: framework}
\end{figure*}

Overall, our key contributions are as follows:
\begin{itemize}[leftmargin=*,itemsep=0pt,parsep=0.5em,topsep=0.3em,partopsep=0.3em]
    \item We introduce \benchmark, a comprehensive benchmark designed to assess the performance of LLMs in data science tasks in Figure~\ref{fig: framework}. We develop a semi-automated pipeline to generate ground truth and evaluate aggregated metrics using carefully crafted complex questions.
    \item We propose an innovative Task-Function-Code (TFC) evaluation framework based on predefined aggregated metrics and programmatic rules. We then assess \textbf{23} large language models from both coarse-grained and fine-grained perspectives, presenting the results in Table~\ref{tab:overall_results}.
    \item Various analyses of comparisons and correlations between ours and existing benchmarks are performed in Figure~\ref{fig: ours_vs_humaneval} and Table~\ref{tab: correlation}. Furthermore, we provide research insights derived from experimental outcomes of the evaluated LLMs that point to interesting directions for future work.
\end{itemize}

\section{Background on using LLMs for Data Science}
This section discusses the key aspects that underlie our benchmarking approach.

\noindent \textbf{Ground Truth Generation.} Ground truth serves as the cornerstone for evaluating the performance of LLMs in data science tasks. For diverse and challenging data science prompts, we aim to propose a semi-automated pipeline that leverages a robust LLM to generate GTs and employs self-consistency and human validation strategies to ensure the accuracy and reliability of generated GTs.

\noindent \textbf{Evaluation Metric Definition.} Defining appropriate and meaningful evaluation metrics is essential for effectively comparing and analyzing the effectiveness of different LLMs in data science tasks. Our study meticulously defines evaluation metrics tailored to the specific tasks and challenges the collected prompts pose. These metrics are designed to capture the diverse nuances of data analysis and visualization tasks, enabling a comprehensive assessment of LLMs' capabilities.

\noindent \textbf{Limitation of Previous Studies.} Prior research in benchmarking LLMs for data science has often been limited by focusing on single tasks, simplistic evaluation metrics, and readily available ground truth. These shortcomings hinder the thorough evaluation of LLMs and may not fully capture their strengths and weaknesses. By addressing these limitations, our study seeks to provide a more comprehensive and nuanced assessment of LLMs in data science. Through the development of \benchmark$\space$ and the implementation of a rigorous evaluation framework, we aim to push the boundaries of benchmarking practices in the field of data science and LLM research.

\section{DataSciBench}
\label{sec: DataSciBench}

\benchmark$\space$ consists of three important components as outlined in Figure~\ref{fig: framework}. 

\begin{itemize}[leftmargin=*,itemsep=0pt,parsep=0.5em,topsep=0.3em,partopsep=0.3em]
    \item \textbf{Prompt Definition and Collection} which defines \textbf{6} task types and collects \textbf{222} real, challenging, and high-quality prompts through question filtering and expert review.

    \item \textbf{Response Integration and Validation} which proposes a novel Task - Function - Code (TFC) that produces \textbf{519} test cases to effectively assess the key tasks of each prompt through defined aggregate functions and programmatic rules.

    \item \textbf{LLM Evaluation} which assesses \textbf{6} API-based models, \textbf{8} open-sourced general models, and \textbf{9} open-sourced code generation models from coarse-grained (i.e. success rate, completion rate) and fine-grained (e.g., Vision-Language Model (VLM)-as-a-judge, \textbf{25} aggregate functions) perspectives.
\end{itemize}

\subsection{Prompt Definition for Data Science}

\vpara{Task Type.} We define six typical data science tasks in Appendix~\ref{ssec: task_des} that include Data cleaning and preprocessing, Data exploration and statistics understanding, Data visualization, Predictive modeling, Data mining and Pattern recognition, and Interpretability and Report generation.

\vpara{Task Integration.} To increase the difficulty of prompts, we chose more complex prompts that included multiple defined task types. These sequential tasks can be any combination of six task types.

\subsection{Dataset Collection}
\vpara{Question Collection.}
We collect questions from four sources:
\textbf{1) Extensive collection from a real-world online platform.} We collect natural prompts from one online code-generation platform, CodeGeeX~\citep{zheng2023codegeex}.
\textbf{2) Extracted and rewritten from a public code benchmark.} We utilize \textbf{167} high-quality data science prompts from BigCodeBench (BCB) and then refine them to our specified format encompassing \textit{input data or file, prompt, and expected output file} with TFC for standardized evaluation.
\textbf{3) Hand-written by humans.} We also write elaborated prompts to increase the difficulty and robustness of the evaluated benchmark by referring to relative websites\footnote{https://ds100.org/course-notes/eda/eda.html}. 
\textbf{4) Synthesized from LLMs.} We use a few shot examples drawn from human-written prompts to ask LLM to generate prompts.

\vpara{Question Filtering.}
We filter low-quality questions via the following principles:
\textbf{1) Choose questions} that include keywords, but are not limited to, ``machine learning'', ``deep learning'', ``data preprocessing'', and ``data visualization'';
\textbf{2) Filter questions} that require rewriting code, finding errors, and explaining basic concepts.

\vpara{Expert Review.}
We review the prompts we collect with experts in computer science and data analysis to ensure their quality.
The review process includes three stages:
\textbf{1) In stage 1,} experts verify the correctness and adjust the suitability of prompts. In addition, experts ensure that the responses to the questions are clear and structured in a way that facilitates assessment. For example, handling missing values in a data frame.
\textbf{2) In stage 2}, experts format all verified prompts into a unified instruction that encompasses \textit{input data or file, prompt, and expected output file}.
\textbf{3) In stage 3}, experts ensure the availability of input prompt datasets by collecting public datasets or generating random datasets.

\begin{figure}[t!]
    \centering
    \includegraphics[width=0.8\linewidth]{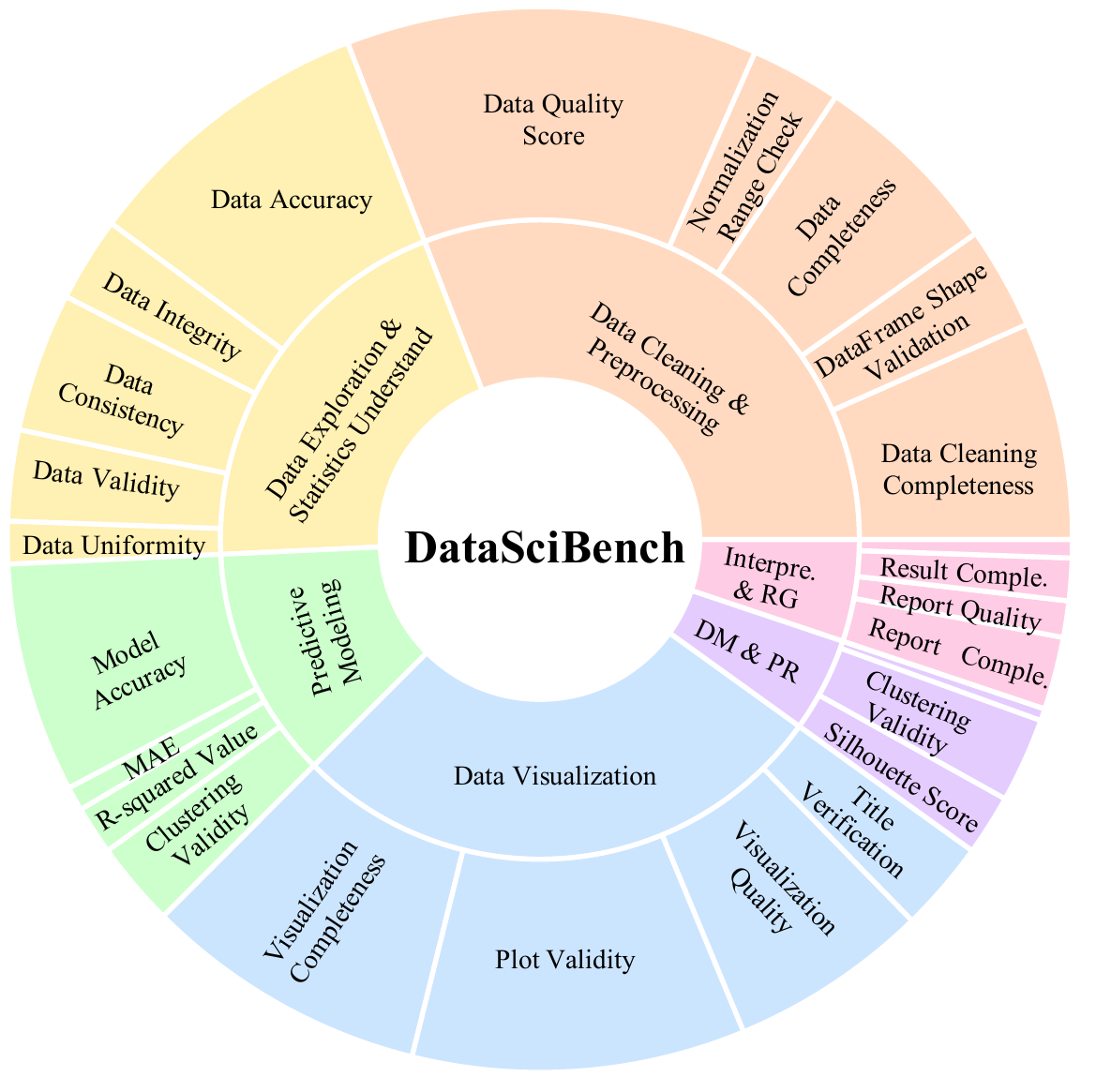}
    \caption{Statistics of \textbf{6} task types and \textbf{25} aggregate functions in Task-Function-Code (TFC) list. DM \& PR denotes Data Mining \& Pattern Recognition. Interpre. \& RG denotes Interpretability \& Report Generation.}
    \label{fig: function}
    \vspace{-0.4cm}
\end{figure}

\subsection{Response Integration and Validation}

\vpara{Ground Truth Generation and Verification.} To obtain the response of the collected questions, we propose the following strategy to generate test cases for each question. 
Firstly, we generate the outputs of each prompt by sampling LLMs several times and then execute the generated code to obtain the final output.
Then, we use two different validation methods to determine whether the LLM-generated answer aligns with the meaning specified in the aggregate functions to ensure the reliability of the answer. 
For questions originating from BCB, where reliable test cases are provided, we validate the generated answer by performing all test cases. Answers that pass all test cases are re-checked by humans and finally considered ground truth. 
As for other prompts, we initially adopt a self-consistency strategy~\citep{wang2022self} to obtain outputs and then ensure their reliability and precision by having six authors of the paper verify the default assigned prompts and corresponding ground truth elements, including task type, evaluation function, programmatic code, and final outputs. In cases where uncertainties arise in the generated outputs, we cross-validate them among three authors.

\vpara{Evaluation Selection.} We introduce a structured approach to identify and evaluate key tasks across six established types.
We first use GPT-4o-mini to select several valuable task types, return the corresponding evaluation functions, and generate the evaluation codes for each prompt to effectively evaluate the capabilities of LLM and reduce the evaluation cost.
Each group data is simplified as a tuple (T, F, C) in generated $\mathbf{R}$ as follows:
\begin{equation}
    \vspace{-0.1cm}
    \mathbf{R} = \{(\text{T}_i, \text{F}_i, \text{C}_i)|_i^N\},
    \vspace{-0.1cm}
\end{equation}
where $N$ is the number of valuable task types per prompt, and this value is different for each question.
Then we use the data interpreter (DI)~\citep{hong2024datainterpreter} to generate a hierarchical directed acyclic graph (DAG) for each prompt, in which each task type is defined as a node at one level in a DAG, as illustrated in Figure~\ref{fig: framework}.
Based on the generated graphs, we take a powerful LLM as a backbone and run all evaluation functions to obtain the ground truth of each task type.
To some extent, this way of verification can avoid the commonly used LLM-as-a-Judge black-box assessment.

\vpara{Function Aggregation.} To unify the key functions and improve the scalability of the evaluation, we aggregate all generated functions to the top-$K$ function category, select top-$K$ functions for each task type, and finally obtain 25 functions, as shown in Figure~\ref{fig: function}.
Generally, $K$ is set as 5. For example, the function category for data cleaning and preprocessing includes Data Cleaning Completeness, DataFrame Shape Validation, Data Completeness, Normalization Range Check, and Data Quality Score.

\vpara{Programmatic Rules.} Regarding aggregate functions with corresponding codes, we define unified rules to verify generated code.
Specifically, we unify all initial outputs as boolean or decimal types ranging between $0$ and $1$. Then, we obtain the final boolean value by comparing ground truth with prediction output depending on the specific task description of aggregate functions.
For example, regarding Data Cleaning Completeness, which calculates the final number of rows/columns after preprocessing, the final output is $1$ if the number is the same as the number of ground truths; otherwise, it is $0$.
For some specific tasks whose output type is decimal, we also set a corresponding threshold to transform the output to boolean for simplicity. For example, the threshold is set to $0.5$ if the aggregate function is silhouette score for data mining and pattern recognition.

\vpara{Summary.} Based on the above three submodules, we obtain \textbf{222} effective prompts and \textbf{519} corresponding test cases for each prompt, \textbf{25} aggregated functions, which help evaluations of \textbf{6} API-based and \textbf{17} open-sourced models.

\section{Experiments}
\label{sec: experiments}

\subsection{Settings}
We construct a comprehensive benchmark on our collected prompts to assess the performance of \textbf{23} different models (e.g., API-based models and open-sourced general/code generation models). 

\begin{itemize}[leftmargin=*,itemsep=0pt,parsep=0.5em,topsep=0.3em,partopsep=0.3em]
    \item \textbf{Six API-based models} include o1-mini~\citep{jaech2024openai}, GPT-4o-mini~\cite{hurst2024gpt4o}, GPT-4o-2024-05-13, GPT-4-Turbo~\citep{achiam2023gpt}, Claude-3.5-Sonnet\footnote{\textcolor{teal}{https://www.anthropic.com/news/claude-3-5-sonnet}}, and GLM-4-Flash~\citep{glm2024chatglm}. 
    \item \textbf{Eight open-sourced general models} include Llama3.1-8B-Instruct, Llama3-8B-Instruct, Qwen2.5-7B-Instruct, Qwen2-1.5/7B-Instruct~\citep{yang2024qwen2}, Gemma2-9B-it~\citep{team2024gemma}, GLM-4-9B-chat, and Yi-1.5-9B-chat-16k~\citep{young2024yi}.
    \item \textbf{Nine open-sourced code models} include Deepseek-Coder-1.3/6.7/33B-Instruct~\citep{guo2024deepseek}, CodeLlama-7/13/34B-Instruct~\citep{roziere2023code}, Qwen2.5-Coder-1.5/7B-Instruct~\citep{hui2024qwen2}, and StarCoder2-15B~\citep{lozhkov2024StarCoder}.
\end{itemize}

\begin{table*}[t!]
    \centering
    \vspace{-0.2cm}
    \caption{Overall evaluation results for \benchmark$\space$ on all our curated prompts.}
    \vspace{-0.2cm}
    \label{tab:overall_results}
    \resizebox{0.97\textwidth}{!}{
    \begin{tabular}{c|c|c|c|c|c|c|c|c|c|c}
    \specialrule{.16em}{0pt}{.65ex}
        \multirow{2}{*}{Models} & \multirow{2}{*}{Size} & \multicolumn{2}{c|}{Coarse-grained Metrics} & \multicolumn{6}{c|}{Fine-grained Metrics} & \multirow{2}{*}{Score}
        \\
        \cmidrule{3-10}
        & & SR (\%) & CR (\%) & VLM & F1 & F2 & F3 & F4 & F5 & \\
        \specialrule{.10em}{.4ex}{.65ex}
        o1-mini                      & N/A  & 29.77 & 45.26 & 2.87 & 44.63 & 19.27 & 36.01 & 30.94 & 23.81 & 38.78\\
        GPT-4o-2024-05-13            & N/A  & \textbf{66.31} & \textbf{68.44} & \textbf{3.91} & \textbf{75.93} & \textbf{56.14} & \textbf{69.33} & \textbf{71.35} & \textbf{57.67} & \textbf{64.51}\\
        GPT-4o-mini                  & N/A  & 50.63 & 57.78 & 3.05 & 60.30 & \underline{48.02} & 57.84 & 59.24 & \underline{53.54} & 54.18\\
        GPT-4-Turbo                  & N/A  & 51.93 & 58.87 & \underline{3.09} & 62.30 & 41.62 & 57.75 & 60.25 & 50.75 & 54.65\\
        Claude-3-5-Sonnet-20240620   & N/A  & 47.48 & 58.11 & 2.14 & 49.07 & 36.94 & 55.84 & 52.87 & 46.04 & 52.29\\
        GLM-4-Flash                  & N/A  & 30.32 & 34.04 & 1.33 & 36.53 & 29.42 & 32.57 & 27.64 & 14.44 & 30.74\\
        \specialrule{.10em}{.4ex}{.65ex}
        Meta-Llama-3.1-8B-Instruct   & 8B   & 24.73 & 33.89 & 1.29 & 38.24 & 18.25 & 21.98 & 22.89 & 25.85 & 29.69\\
        Meta-Llama-3-8B-Instruct     & 8B   & 2.88  & 3.92  & 0.52 & 4.18  & 1.26  & 2.70  & 2.67  & 1.47  & 3.33\\
        Gemma-2-9B-it                & 9B   & 7.07  & 11.00 & 1.06 & 26.16 & 16.90 & 23.81 & 18.11 & 17.15 & 12.66\\
        GLM-4-9B-Chat                & 9B   & 25.72 & 30.38 & 1.69 & 31.51 & 23.15 & 28.07 & 27.19 & 19.14 & 27.57\\
        Qwen2.5-7B-Instruct          & 7B   & 43.83 & 50.74 & 1.43 & 51.18 & 36.41 & 47.25 & 45.24 & 34.77 & 45.99\\
        Qwen2-7B-Instruct            & 7B   & 22.84 & 25.58 & 1.16 & 30.93 & 20.78 & 28.73 & 25.87 & 7.52  & 23.52\\
        Qwen2-1.5B-Instruct          & 1.5B & 3.96  & 5.46  & 0.40 & 4.54  & 1.98  & 3.26  & 5.76  & 4.71  & 4.78\\
        Yi-1.5-9B-Chat-16K           & 9B   & 38.20 & 42.35 & 0.73 & 38.14 & 36.36 & 35.64 & 37.08 & 27.79 & 38.22\\
        \specialrule{.10em}{.4ex}{.65ex}
        CodeLlama-34B-Instruct    & 34B  & 0.90  & 1.47  & \text{0.00} & 1.02  & 0.84  & 1.98  & 1.54  & 1.19  & 1.33\\
        CodeLlama-13B-Instruct    & 13B  & 10.49 & 14.64 & 0.04 & 11.67 & 11.34 & 9.43  & 14.43 & 5.15  & 12.64\\
        CodeLlama-7B-Instruct     & 7B   & 2.88  & 3.97  & \text{0.00} & 3.53  & 2.37  & 2.57  & 1.74  & 1.59  & 3.31\\
        StarCoder2-15B               & 15B  & 2.07  & 2.61  & 0.07 & 2.57  & 1.81  & 1.59  & 3.43  & 1.19  & 2.33\\
        Deepseek-Coder-33B-instruct  & 33B  & \underline{55.86} & \underline{61.23} & 2.29 & \underline{65.66} & 47.11 & \underline{58.17} & \underline{61.65} & 48.60 & \underline{56.76}\\
        Deepseek-Coder-6.7B-instruct & 6.7B & 37.03 & 41.62 & 1.93 & 43.49 & 34.57 & 46.36 & 46.49 & 18.09 & 38.45\\
        Deepseek-Coder-1.3B-instruct & 1.3B & 15.50 & 19.00 & 0.10 & 13.04 & 14.62 & 13.26 & 16.32 & 7.92  & 16.39\\
        Qwen2.5-Coder-7B-Instruct    & 7B   & 45.18 & 53.11 & 1.48 & 51.58 & 43.21 & 43.87 & 42.50 & 35.23 & 47.67\\
        Qwen2.5-Coder-1.5B-Instruct  & 1.5B & 22.74 & 28.64 & 0.81 & 29.82 & 21.79 & 23.96 & 29.58 & 16.39 & 25.87\\
        \specialrule{.16em}{.4ex}{0pt}
    \end{tabular}
    }
\end{table*}

\subsection{Evaluation Metrics}
\vpara{Coarse-Grained Metrics.} We define the coarse-grained metrics (CR and SR) for evaluating LLMs.

{\noindent \bf $\bullet$ Completion Rate (CR).} Following the CR in Data Interpreter \citep{hong2024datainterpreter}, we calculate the CR  for each TFC in $\mathbf{R}$.
Specifically, we give it a completion score, with a minimum score of $0$ and a maximum score of $2$. The step completion scores were given as follows: missing (score of $0$), fail (score of $0$), success-non-compliant (score of $1$), and success-compliant (score of $2$). The completion rate is then calculated as follows:
\begin{equation}
    \vspace{-0.1cm}
    \text{Completion Rate (CR)} = \frac{\sum_{t=1}^{T} s_t}{T \times s_{\text{max}}},
    \vspace{-0.1cm}
    \label{eq:completion_rate}
\end{equation}
where the numerator was the sum of the completion scores for each step, and the denominator was the sum of the maximum possible scores for all steps ($2 \times T$ and $T$ is the number of TFCs).

{\noindent \bf $\bullet$ Success Rate (SR).} Similar to Codex~\citep{chen2021evaluating}, our success rate is defined as the rate of complete success on a single prompt estimated under 10 runs. Specifically, if all the TFC in $\mathbf{R}$ have passed within a run of a single prompt, it will count as a success. Otherwise, it will count as a failure. Note that for prompts acquired from BigCodeBench, we compare the completion function's outputs with the ground truth completion function's outputs to determine whether a single run passes, since $\mathbf{R}$ is derived based on demanded function outputs in this case. The formula for calculating SR is as follows:
\begin{equation}
    \vspace{-0.1cm}
    \text{Success Rate (SR)} \coloneqq  \underset{\text{Prompts}}{\mathbb{E}} \left[ 1 - \frac{\binom{n-c}{k}}{\binom{n}{k}} \right],
    \vspace{-0.1cm}
\end{equation}
where $n=10$ and $k=1$ in our case, $c$ denotes the number of runs that have passed all TFCs in $\mathbf{R}$.

\vpara{Fine-grained Aggregate Metrics.} We also define the fine-grained aggregate metrics for detail evaluating all LLMs, as shown in Figure~\ref{fig: function} and Table~\ref{tab: program_rules}. 
It's worth noting that data visualization is one of the more challenging tasks. It confirms the existence of outputs but also validates their accuracy and completeness.
Specifically, for tasks like data visualization, we focus on metrics such as visualization quality and use VLM-as-a-judge to assess the quality of images. Finally, we select six representative aggregated metrics as follows:

\begin{itemize}[leftmargin=*,itemsep=0pt,parsep=0.5em,topsep=0.3em,partopsep=0.3em]
    \item VLM-as-a-judge assesses the overall score of two visual inputs based on predefined criteria for progressively adding scores (see Appendix~\ref{sec:vlm_prompt}) to write the total score and provide a brief step-by-step reasoning explanation for its evaluation, with GPT-4o-mini serving as the VLM.
    \item Data Quality Score (F1) in Data cleaning and preprocessing aims to assess the cleanliness of data post-processing. It yields a boolean output of $1$ if it matches the ground truth or $0$ otherwise.
    \item Plot Validity (F2) in Data visualization pertains to the accuracy of visual representations, such as checking whether the shape of an association matrix is consistent with the ground truth. If consistent, the final value is $1$, otherwise $0$.
    \item Data Accuracy (F3) in Data Exploration and Statistics Understand focuses on understanding data quality and can be quantified using Mean Squared Error (MSE). The final value is derived by comparing it against the ground truth with a predefined threshold.
    \item Visualization Completeness (F4) in Data visualization evaluates the comprehensiveness of generated images (e.g., PNG, jpeg, PDF) by checking their existence compared to the ground truth. A score of $1$ is assigned if the files exist, and $0$ otherwise. Note that we use VLM-as-a-judge and F2 to assess the quality of generated images. 
    \item Model Accuracy (F5) in Predictive modeling is utilized to gauge the predictive performance of models, providing a boolean accuracy value or decimal ranging between $0$ and $1$.
\end{itemize}

\vpara{Final Score.} We calculate the final score through the following definition:
\small
\begin{equation}
    \vspace{-0.1cm}
    \text{Score}=0.05 \times \sum_{i=1}^{5}\text{F}_i + 0.05 \times \text{S}_\text{VLM} + 0.05 \times \text{SR} + 0.65 \times \text{CR},
    \vspace{-0.05cm}
\end{equation}
\normalsize
here, the final Score is computed by combining the weighted sum of five metrics $\text{F}_{i}$ (where i ranges from $1$ to $5$), a value VLM score, a value SR, and a value CR. Each factor is multiplied by its respective weight (0.05 or 0.65).

\begin{figure}[t!]
    \centering
    \includegraphics[width=0.98\linewidth]{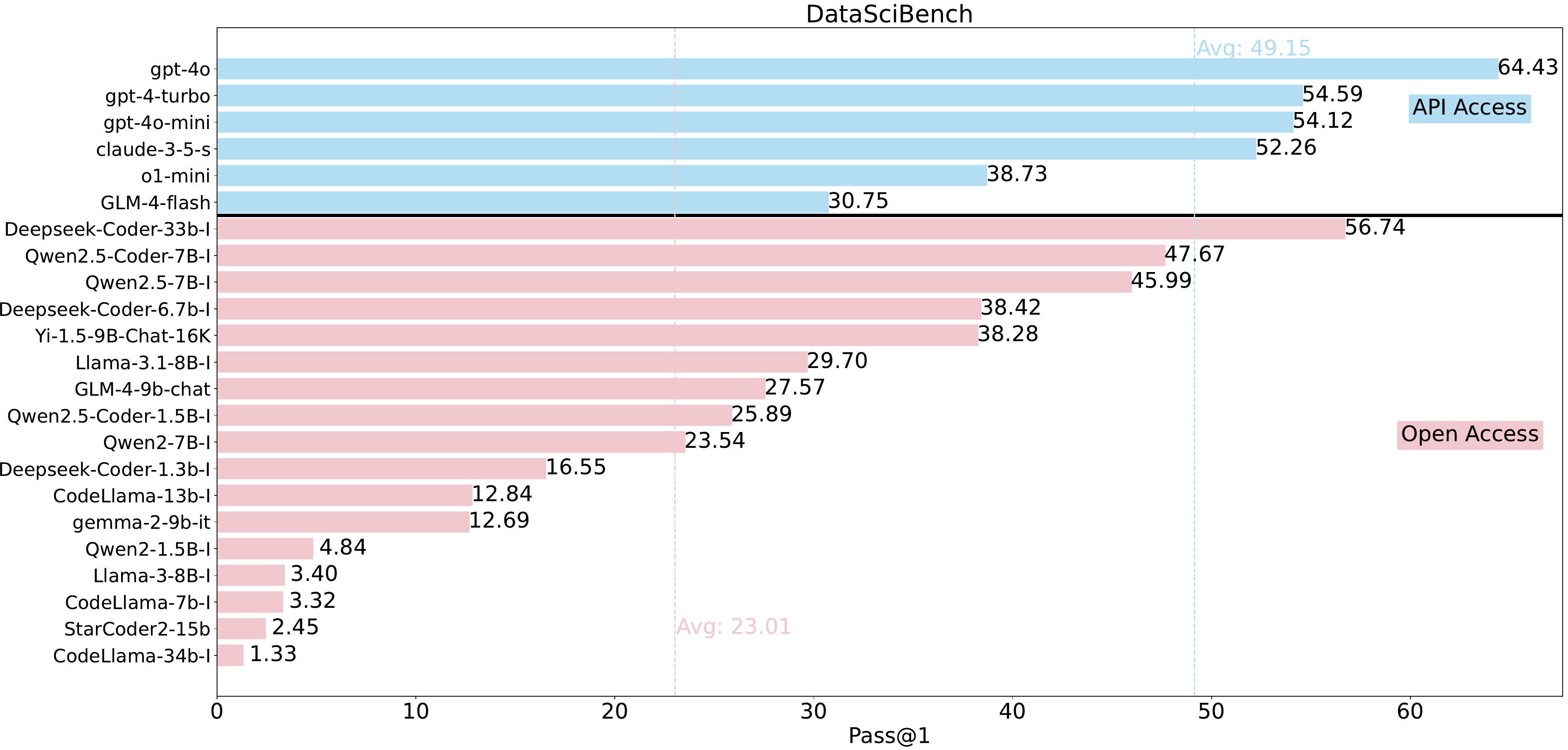}
    \vspace{-0.3cm}
    \caption{Overall score results of all tested LLMs.}
    \vspace{-0.3cm}
    \label{fig: final_evaluation}
\end{figure}

\section{Results and Analysis}

\subsection{Overall Performance}

We demonstrate overall experiment results in Table~\ref{tab:overall_results} and Figure~\ref{fig: final_evaluation}. The key findings are: \textbf{1) Concerning average performance, API-based models outperform open-sourced models.} Among API-based models, GPT-4o achieves the highest total score of $64.51\%$, attaining a significant $9.86\%$ advantage over GPT-4-Turbo, which achieves $54.65\%$ total score. GPT-4o also surpasses all other models on all metrics, indicating its comprehensive capacity over various aspects. 
In comparison, the performance variance between API-based models is smaller than that of open-sourced models.
\textbf{2) As for open-sourced models, the performance gap between general models and code generation models is insignificant.} Among those, Deepseek-Coder-33B-Instruct achieves the highest score of $56.76\%$, even outperforming various API-based models like o1-mini and GPT-4-Turbo. Other models like Qwen2.5-Coder-7B-Instruct and Qwen2.5-7B-Instruct also show fair good capability, attaining total scores of $47.67\%$ and $45.99\%$, respectively. In contrast, a few models only pass very few tasks, achieving total scores even lower than $5.0\%$. Of these, CodeLlama-34B-Instruct unexpectedly achieves a score of $1.33\%$, even lagging behind its small-scale version CodeLlama-7B-Instruct. In this regard, we also present an analysis of the anomaly in Section~\ref{sec: insights}.

\subsection{Study with Different Difficulty Levels}
\begin{figure}[t!]
    \centering
    \includegraphics[width=0.98\linewidth]{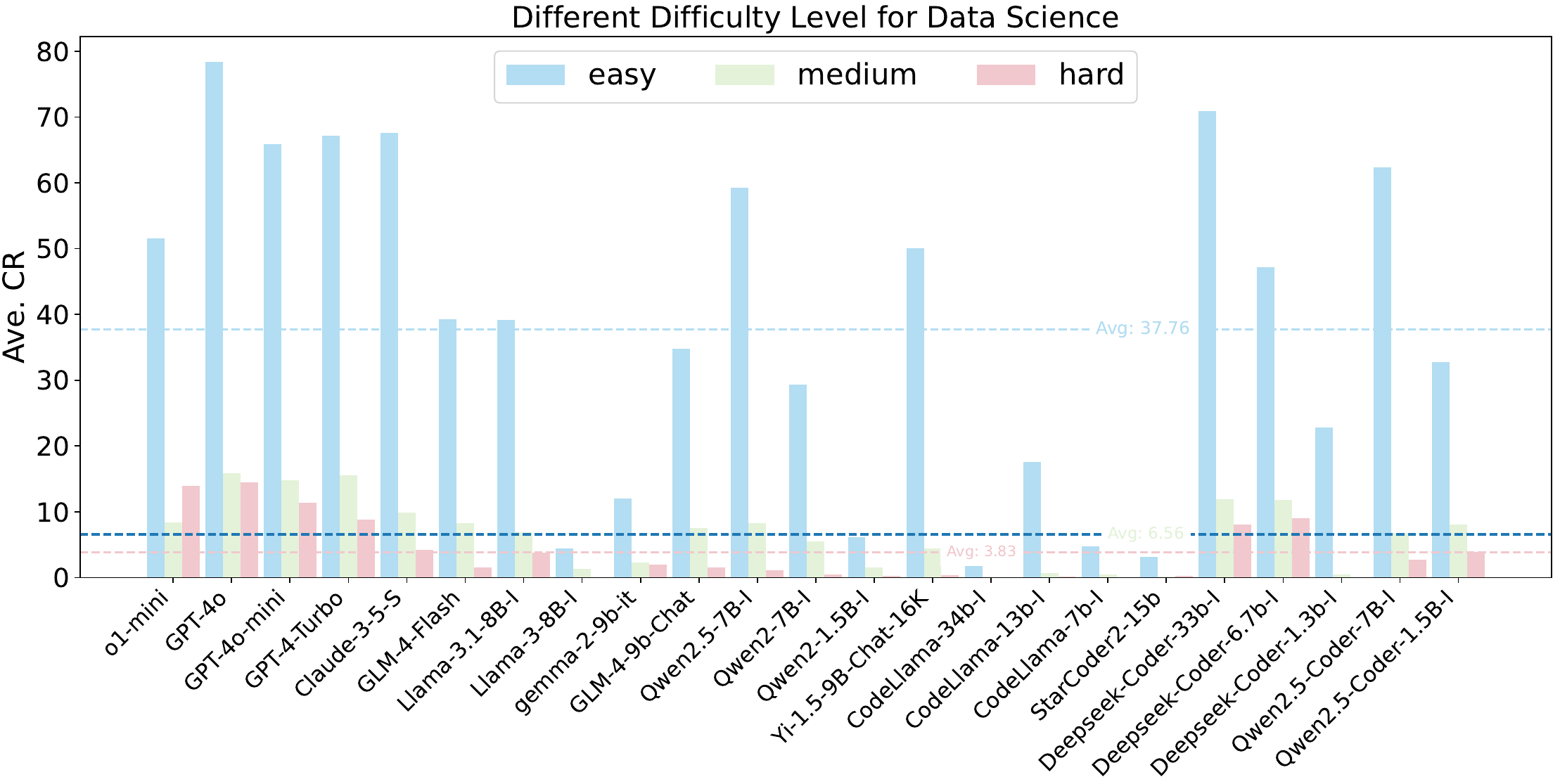}
    \vspace{-0.3cm}
    \caption{Average Completion Rate results regarding different difficulty levels.}
    \vspace{-0.3cm}
    \label{fig: task_level}
\end{figure}
To evaluate multiple LLMs on their ability to complete prompts of varying difficulty, we categorized tasks using BCB and data formatted in CSV, human handwritten prompts, and data science-related DL tasks as easy - 167, medium - 30, and hard levels - 25, respectively. 
We assessed multiple LLMs by combining different difficulty levels, overall average CR, and the average CR for each difficulty level. From the Figure~\ref{fig: task_level}, it can be observed that: 
\textbf{1)} Some LLMs, like GPT-4o, GPT-4o-mini, GPT-4-Turbo, and Deepseek-Coder-33B-Instruct, exhibit consistent performance across all difficulty levels, indicating robustness.
\textbf{2)} Models such as GPT-4 series and Deepseek-Coder-Instruct series are among the top performers, scoring high average CRs, particularly excelling in complex, data-driven tasks defined as hard.
\textbf{3)} There are noticeable performance disparities among general models and small-scale models in terms of average metrics, such CR, with some achieving lower scores overall, suggesting that general models are less efficient or accurate in data science tasks.

\begin{figure}[t!]
    \centering
\includegraphics[width=0.98\linewidth]{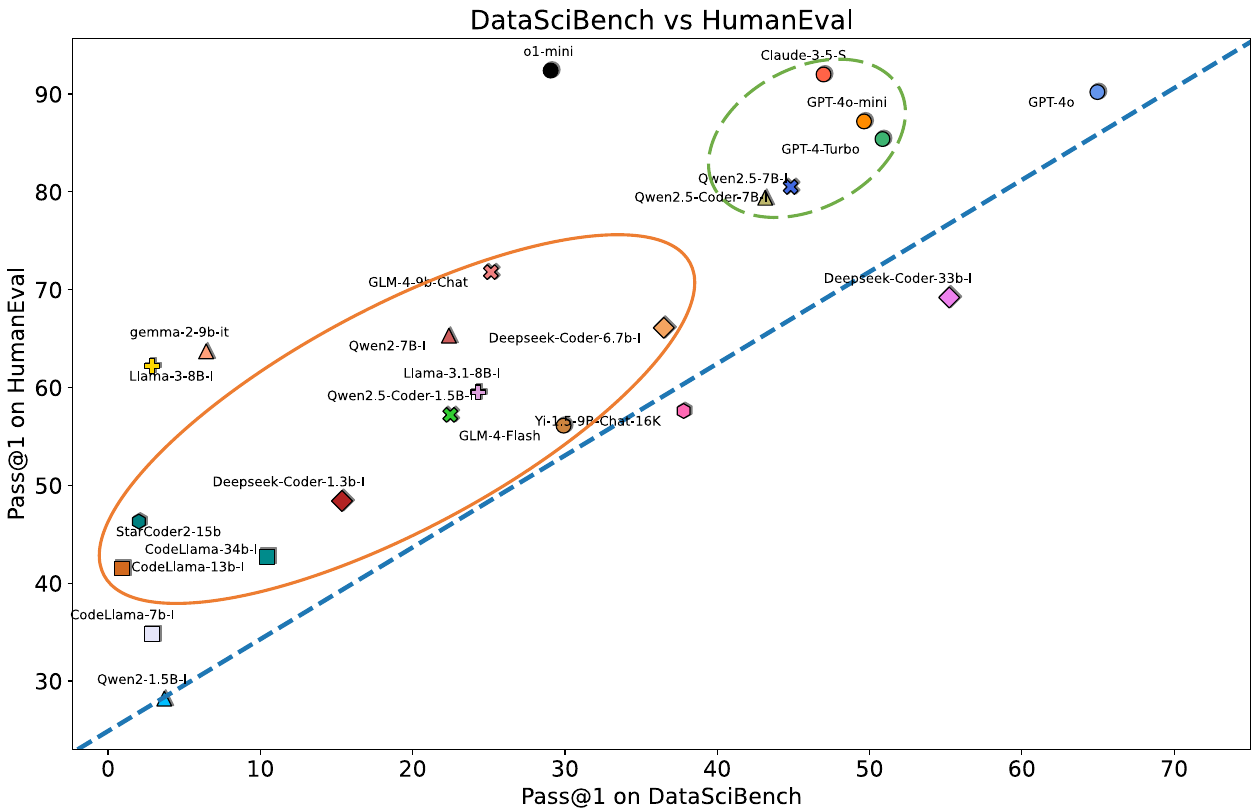}
    \vspace{-0.1cm}
    \caption{Pass@1 comparison of all tested LLMs between \benchmark$\space$ and HumanEval. Circle markers denote the API-based models while others denote various open-sourced LLMs. The green dashed areas indicate the LLMs perform well on the two benchmarks and the orange solid areas indicate performances of the two datasets are relatively mismatched. 
    }
    \label{fig: ours_vs_humaneval}
    \vspace{-0.1cm}
\end{figure}

\subsection{Comparison with HumanEval}
We compare our proposed \benchmark$\space$ with HumanEval.
As shown in Figure~\ref{fig: ours_vs_humaneval}, we observe that most LLMs are located in the upper triangular region of the graph and all tested models are divided into two groups, 
in which the green-dashed-line areas where LLMs perform well on the two benchmarks and the orange-solid-line area where performances on the two datasets with the same model indicate significant discrepancies.

\subsection{Deeper Insights into LLMs' Ability}
\label{sec: insights}
With curated metrics, we can obtain deeper research insights into LLMs' ability to plan and execute complex data science tasks. The experiment results also raise questions that are worth exploring since some results do not conform to conventional perceptions.

\vpara{Models demonstrate proficiency in reasoning tasks but may not consistently excel in complex data science tasks that necessitate precise adherence to detailed instructions, utilization of existing tools, and strategic planning.}
Although, indeed, data science coding tasks often involve scheduling and step-by-step execution similar to reasoning scenarios, results show that even the LLMs proficient in reasoning tasks can still fail to complete complex data science tasks. For instance, the o1-mini model, which is commonly regarded as one of the best reasoning models, unexpectedly failed on many of \benchmark's tasks. The model only achieves a 29.77\% overall success rate, significantly lagging behind the company's previously introduced models like GPT-4o and GPT-4-Turbo. After examining the completions generated by o1-mini, we discovered that the failures are primarily caused by non-compliance with instructions, incorrect calls, and forgetfulness. While successfully splitting the task into multiple subtasks, the model often forgets to export required execution outcomes or just outputs undesired data. In other cases, the model may falsely call a library function or method that sometimes does not even exist. 
The main kind of coding error is an execution error. An example is a hallucination about the column name of a CSV file.
These facts remind us that real-life data science coding tasks often comprehensively challenge the model's ability to follow fine-grained instructions, utilize existing tools (libraries, APIs$\ldots$), and do planning. To excel and align effectively with these tasks, a model must demonstrate competitiveness across all relevant aspects.
We included four bad examples generated by o1-mini for further clarification in Appendix~\ref{appendix: worse_example}.

\vpara{Larger-scale models may encounter challenges in following simple instructions due to the extensive data utilized during training, potentially leading to difficulties in generating outputs in formats different from those within the training data.}
StarCoder2-15B performs worse than some smaller models, and CodeLlama-34B-Instruct even performs worse than 13B and 7B versions. The main reason is that the larger-scale version lacks some other ability like generating formatted text according to prompts. Perhaps a large amount of data in a certain format is being used to train a larger version that fails to follow the prompt to generate another format different from that. We present some examples in Appendix~\ref{appendix: worse_example}. Indeed, the larger-scale model of CodeLlama also fails to outperform the smaller-scale version in LCB.

\section{Conclusion}
\vspace{-0.2cm}
\label{sec: discussion}
This paper introduces \benchmark, a novel framework tailored to assess the capabilities of LLMs in data science tasks. By meticulously curating challenging prompts and leveraging robust LLMs alongside a self-consistency strategy, we generate ground truth for all prompts. To efficiently evaluate LLMs' performance, we aggregate evaluation metrics and synthesize the TFC list programmatically. Subsequently, we assess \textbf{23} API-based and open-source models, offer valuable research and engineering insights, and present error analyses of the assessed LLMs.

\section{Limitations}
In certain visualization tasks, our initial metrics and evaluation methods (e.g., VLM-as-a-judge) may lack precision. Further refinement of metrics is required to evaluate data visualization tasks effectively. One potential approach could involve employing VLMs to train critic models, enhancing the capability for fine-grained evaluations of visualizations.




\bibliography{ref}

\begin{thebibliography}{36}
\expandafter\ifx\csname natexlab\endcsname\relax\def\natexlab#1{#1}\fi

\bibitem[{Achiam et~al.(2023)Achiam, Adler, Agarwal, Ahmad, Akkaya, Aleman, Almeida, Altenschmidt, Altman, Anadkat et~al.}]{achiam2023gpt}
Josh Achiam, Steven Adler, Sandhini Agarwal, Lama Ahmad, Ilge Akkaya, Florencia~Leoni Aleman, Diogo Almeida, Janko Altenschmidt, Sam Altman, Shyamal Anadkat, et~al. 2023.
\newblock Gpt-4 technical report.
\newblock \emph{arXiv preprint arXiv:2303.08774}.

\bibitem[{Chen et~al.(2024{\natexlab{a}})Chen, Cheng, Li, Geng, Gong, Li, Bei, Tan, Wang, Zeng et~al.}]{chen2024xtrimopglm}
Bo~Chen, Xingyi Cheng, Pan Li, Yangli-ao Geng, Jing Gong, Shen Li, Zhilei Bei, Xu~Tan, Boyan Wang, Xin Zeng, et~al. 2024{\natexlab{a}}.
\newblock xtrimopglm: unified 100b-scale pre-trained transformer for deciphering the language of protein.
\newblock \emph{arXiv preprint arXiv:2401.06199}.

\bibitem[{Chen et~al.(2021)Chen, Tworek, Jun, Yuan, Pinto, Kaplan, Edwards, Burda, Joseph, Brockman et~al.}]{chen2021evaluating}
Mark Chen, Jerry Tworek, Heewoo Jun, Qiming Yuan, Henrique Ponde De~Oliveira Pinto, Jared Kaplan, Harri Edwards, Yuri Burda, Nicholas Joseph, Greg Brockman, et~al. 2021.
\newblock Evaluating large language models trained on code.
\newblock \emph{arXiv preprint arXiv:2107.03374}.

\bibitem[{Chen et~al.(2024{\natexlab{b}})Chen, Zhang, Xu, Ren, and Yang}]{chen2024viseval}
Nan Chen, Yuge Zhang, Jiahang Xu, Kan Ren, and Yuqing Yang. 2024{\natexlab{b}}.
\newblock Viseval: A benchmark for data visualization in the era of large language models.
\newblock \emph{IEEE Transactions on Visualization and Computer Graphics}.

\bibitem[{GLM et~al.(2024)GLM, Zeng, Xu, Wang, Zhang, Yin, Rojas, Feng, Zhao, Lai et~al.}]{glm2024chatglm}
Team GLM, Aohan Zeng, Bin Xu, Bowen Wang, Chenhui Zhang, Da~Yin, Diego Rojas, Guanyu Feng, Hanlin Zhao, Hanyu Lai, et~al. 2024.
\newblock Chatglm: A family of large language models from glm-130b to glm-4 all tools.
\newblock \emph{arXiv preprint arXiv:2406.12793}.

\bibitem[{Guo et~al.(2024)Guo, Zhu, Yang, Xie, Dong, Zhang, Chen, Bi, Wu, Li et~al.}]{guo2024deepseek}
Daya Guo, Qihao Zhu, Dejian Yang, Zhenda Xie, Kai Dong, Wentao Zhang, Guanting Chen, Xiao Bi, Yu~Wu, YK~Li, et~al. 2024.
\newblock Deepseek-coder: When the large language model meets programming--the rise of code intelligence.
\newblock \emph{arXiv preprint arXiv:2401.14196}.

\bibitem[{He et~al.(2023)He, Zhou, Xu, Ma, Ding, Du, Gao, Jia, Chen, Han, Yuan, and Zhang}]{he2024text2analysis}
Xinyi He, Mengyu Zhou, Xinrun Xu, Xiaojun Ma, Rui Ding, Lun Du, Yan Gao, Ran Jia, Xu~Chen, Shi Han, Zejian Yuan, and Dongmei Zhang. 2023.
\newblock \href {http://arxiv.org/abs/2312.13671} {Text2analysis: A benchmark of table question answering with advanced data analysis and unclear queries}.

\bibitem[{Hong et~al.(2024)Hong, Lin, Liu, Wu, Li, Chen, Zhang, Wang, Zhang, Zhuge et~al.}]{hong2024datainterpreter}
Sirui Hong, Yizhang Lin, Bangbang Liu, Binhao Wu, Danyang Li, Jiaqi Chen, Jiayi Zhang, Jinlin Wang, Lingyao Zhang, Mingchen Zhuge, et~al. 2024.
\newblock Data interpreter: An llm agent for data science.
\newblock \emph{arXiv preprint arXiv:2402.18679}.

\bibitem[{Hu et~al.(2024)Hu, Zhao, Wei, Chai, Ma, Wang, Wang, Su, Xu, Zhu et~al.}]{huinfiagent}
Xueyu Hu, Ziyu Zhao, Shuang Wei, Ziwei Chai, Qianli Ma, Guoyin Wang, Xuwu Wang, Jing Su, Jingjing Xu, Ming Zhu, et~al. 2024.
\newblock Infiagent-dabench: Evaluating agents on data analysis tasks.
\newblock In \emph{Forty-first International Conference on Machine Learning}.

\bibitem[{Huang et~al.(2023)Huang, Vora, Liang, and Leskovec}]{huang2023mlbenchmarking}
Qian Huang, Jian Vora, Percy Liang, and Jure Leskovec. 2023.
\newblock Benchmarking large language models as ai research agents.
\newblock \emph{arXiv preprint arXiv:2310.03302}.

\bibitem[{Hui et~al.(2024)Hui, Yang, Cui, Yang, Liu, Zhang, Liu, Zhang, Yu, Dang et~al.}]{hui2024qwen2}
Binyuan Hui, Jian Yang, Zeyu Cui, Jiaxi Yang, Dayiheng Liu, Lei Zhang, Tianyu Liu, Jiajun Zhang, Bowen Yu, Kai Dang, et~al. 2024.
\newblock Qwen2. 5-coder technical report.
\newblock \emph{arXiv preprint arXiv:2409.12186}.

\bibitem[{Hurst et~al.(2024)Hurst, Lerer, Goucher, Perelman, Ramesh, Clark, Ostrow, Welihinda, Hayes, Radford et~al.}]{hurst2024gpt4o}
Aaron Hurst, Adam Lerer, Adam~P Goucher, Adam Perelman, Aditya Ramesh, Aidan Clark, AJ~Ostrow, Akila Welihinda, Alan Hayes, Alec Radford, et~al. 2024.
\newblock Gpt-4o system card.
\newblock \emph{arXiv preprint arXiv:2410.21276}.

\bibitem[{Jaech et~al.(2024)Jaech, Kalai, Lerer, Richardson, El-Kishky, Low, Helyar, Madry, Beutel, Carney et~al.}]{jaech2024openai}
Aaron Jaech, Adam Kalai, Adam Lerer, Adam Richardson, Ahmed El-Kishky, Aiden Low, Alec Helyar, Aleksander Madry, Alex Beutel, Alex Carney, et~al. 2024.
\newblock Openai o1 system card.
\newblock \emph{arXiv preprint arXiv:2412.16720}.

\bibitem[{Jain et~al.(2024)Jain, Han, Gu, Li, Yan, Zhang, Wang, Solar-Lezama, Sen, and Stoica}]{jain2024livecodebench}
Naman Jain, King Han, Alex Gu, Wen-Ding Li, Fanjia Yan, Tianjun Zhang, Sida Wang, Armando Solar-Lezama, Koushik Sen, and Ion Stoica. 2024.
\newblock Livecodebench: Holistic and contamination free evaluation of large language models for code.
\newblock \emph{arXiv preprint arXiv:2403.07974}.

\bibitem[{Jumper et~al.(2021)Jumper, Evans, Pritzel, Green, Figurnov, Ronneberger, Tunyasuvunakool, Bates, {\v{Z}}{\'\i}dek, Potapenko et~al.}]{jumper2021highly}
John Jumper, Richard Evans, Alexander Pritzel, Tim Green, Michael Figurnov, Olaf Ronneberger, Kathryn Tunyasuvunakool, Russ Bates, Augustin {\v{Z}}{\'\i}dek, Anna Potapenko, et~al. 2021.
\newblock Highly accurate protein structure prediction with alphafold.
\newblock \emph{nature}, 596(7873):583--589.

\bibitem[{Lai et~al.(2023)Lai, Li, Wang, Zhang, Zhong, Zettlemoyer, Yih, Fried, Wang, and Yu}]{lai2023ds}
Yuhang Lai, Chengxi Li, Yiming Wang, Tianyi Zhang, Ruiqi Zhong, Luke Zettlemoyer, Wen-tau Yih, Daniel Fried, Sida Wang, and Tao Yu. 2023.
\newblock Ds-1000: A natural and reliable benchmark for data science code generation.
\newblock In \emph{International Conference on Machine Learning}, pages 18319--18345. PMLR.

\bibitem[{Li et~al.(2024)Li, Su, Chen, Li, and ZHANG}]{li2024sheetcopilot}
Hongxin Li, Jingran Su, Yuntao Chen, Qing Li, and ZHAO-XIANG ZHANG. 2024.
\newblock Sheetcopilot: Bringing software productivity to the next level through large language models.
\newblock \emph{Advances in Neural Information Processing Systems}, 36.

\bibitem[{Lozhkov et~al.(2024)Lozhkov, Li, Allal, Cassano, Lamy-Poirier, Tazi, Tang, Pykhtar, Liu, Wei et~al.}]{lozhkov2024StarCoder}
Anton Lozhkov, Raymond Li, Loubna~Ben Allal, Federico Cassano, Joel Lamy-Poirier, Nouamane Tazi, Ao~Tang, Dmytro Pykhtar, Jiawei Liu, Yuxiang Wei, et~al. 2024.
\newblock Starcoder 2 and the stack v2: The next generation.
\newblock \emph{arXiv preprint arXiv:2402.19173}.

\bibitem[{Lu et~al.(2024)Lu, Lu, Lange, Foerster, Clune, and Ha}]{lu2024ai}
Chris Lu, Cong Lu, Robert~Tjarko Lange, Jakob Foerster, Jeff Clune, and David Ha. 2024.
\newblock The ai scientist: Towards fully automated open-ended scientific discovery.
\newblock \emph{arXiv preprint arXiv:2408.06292}.

\bibitem[{Ma et~al.(2023)Ma, Ding, Wang, Han, and Zhang}]{ma2023insightpilot}
Pingchuan Ma, Rui Ding, Shuai Wang, Shi Han, and Dongmei Zhang. 2023.
\newblock Insightpilot: An llm-empowered automated data exploration system.
\newblock In \emph{Proceedings of the 2023 Conference on Empirical Methods in Natural Language Processing: System Demonstrations}, pages 346--352.

\bibitem[{Roziere et~al.(2023)Roziere, Gehring, Gloeckle, Sootla, Gat, Tan, Adi, Liu, Sauvestre, Remez et~al.}]{roziere2023code}
Baptiste Roziere, Jonas Gehring, Fabian Gloeckle, Sten Sootla, Itai Gat, Xiaoqing~Ellen Tan, Yossi Adi, Jingyu Liu, Romain Sauvestre, Tal Remez, et~al. 2023.
\newblock Code llama: Open foundation models for code.
\newblock \emph{arXiv preprint arXiv:2308.12950}.

\bibitem[{Team et~al.(2023)Team, Anil, Borgeaud, Wu, Alayrac, Yu, Soricut, Schalkwyk, Dai, Hauth et~al.}]{team2023gemini}
Gemini Team, Rohan Anil, Sebastian Borgeaud, Yonghui Wu, Jean-Baptiste Alayrac, Jiahui Yu, Radu Soricut, Johan Schalkwyk, Andrew~M Dai, Anja Hauth, et~al. 2023.
\newblock Gemini: a family of highly capable multimodal models.
\newblock \emph{arXiv preprint arXiv:2312.11805}.

\bibitem[{Team et~al.(2024)Team, Mesnard, Hardin, Dadashi, Bhupatiraju, Pathak, Sifre, Rivi{\`e}re, Kale, Love et~al.}]{team2024gemma}
Gemma Team, Thomas Mesnard, Cassidy Hardin, Robert Dadashi, Surya Bhupatiraju, Shreya Pathak, Laurent Sifre, Morgane Rivi{\`e}re, Mihir~Sanjay Kale, Juliette Love, et~al. 2024.
\newblock Gemma: Open models based on gemini research and technology.
\newblock \emph{arXiv preprint arXiv:2403.08295}.

\bibitem[{Wang et~al.(2024)Wang, Zhang, Feng, Wang, and Tang}]{wang2024battleagentbench}
Wei Wang, Dan Zhang, Tao Feng, Boyan Wang, and Jie Tang. 2024.
\newblock Battleagentbench: A benchmark for evaluating cooperation and competition capabilities of language models in multi-agent systems.
\newblock \emph{arXiv preprint arXiv:2408.15971}.

\bibitem[{Wang et~al.(2022)Wang, Wei, Schuurmans, Le, Chi, Narang, Chowdhery, and Zhou}]{wang2022self}
Xuezhi Wang, Jason Wei, Dale Schuurmans, Quoc Le, Ed~Chi, Sharan Narang, Aakanksha Chowdhery, and Denny Zhou. 2022.
\newblock Self-consistency improves chain of thought reasoning in language models.
\newblock \emph{arXiv preprint arXiv:2203.11171}.

\bibitem[{Xia et~al.(2024)Xia, Zhang, Liao, Hou, Sun, Li, Fu, and Dong}]{xia2024scenegenagent}
Xiao Xia, Dan Zhang, Zibo Liao, Zhenyu Hou, Tianrui Sun, Jing Li, Ling Fu, and Yuxiao Dong. 2024.
\newblock Scenegenagent: Precise industrial scene generation with coding agent.
\newblock \emph{arXiv preprint arXiv:2410.21909}.

\bibitem[{Xu et~al.(2024)Xu, Liu, Sun, Cheng, Yu, Lai, Zhang, Zhang, Tang, and Dong}]{xu2024androidlab}
Yifan Xu, Xiao Liu, Xueqiao Sun, Siyi Cheng, Hao Yu, Hanyu Lai, Shudan Zhang, Dan Zhang, Jie Tang, and Yuxiao Dong. 2024.
\newblock Androidlab: Training and systematic benchmarking of android autonomous agents.
\newblock \emph{arXiv preprint arXiv:2410.24024}.

\bibitem[{Yang et~al.(2024)Yang, Yang, Hui, Zheng, Yu, Zhou, Li, Li, Liu, Huang et~al.}]{yang2024qwen2}
An~Yang, Baosong Yang, Binyuan Hui, Bo~Zheng, Bowen Yu, Chang Zhou, Chengpeng Li, Chengyuan Li, Dayiheng Liu, Fei Huang, et~al. 2024.
\newblock Qwen2 technical report.
\newblock \emph{arXiv preprint arXiv:2407.10671}.

\bibitem[{Yao et~al.(2022)Yao, Zhao, Yu, Du, Shafran, Narasimhan, and Cao}]{yao2022react}
Shunyu Yao, Jeffrey Zhao, Dian Yu, Nan Du, Izhak Shafran, Karthik Narasimhan, and Yuan Cao. 2022.
\newblock React: Synergizing reasoning and acting in language models.
\newblock \emph{arXiv preprint arXiv:2210.03629}.

\bibitem[{Young et~al.(2024)Young, Chen, Li, Huang, Zhang, Zhang, Li, Zhu, Chen, Chang et~al.}]{young2024yi}
Alex Young, Bei Chen, Chao Li, Chengen Huang, Ge~Zhang, Guanwei Zhang, Heng Li, Jiangcheng Zhu, Jianqun Chen, Jing Chang, et~al. 2024.
\newblock Yi: Open foundation models by 01. ai.
\newblock \emph{arXiv preprint arXiv:2403.04652}.

\bibitem[{Zhang et~al.(2024{\natexlab{a}})Zhang, Hu, Zhoubian, Du, Yang, Wang, Yue, Dong, and Tang}]{zhang2024sciglm}
Dan Zhang, Ziniu Hu, Sining Zhoubian, Zhengxiao Du, Kaiyu Yang, Zihan Wang, Yisong Yue, Yuxiao Dong, and Jie Tang. 2024{\natexlab{a}}.
\newblock Sciglm: Training scientific language models with self-reflective instruction annotation and tuning.
\newblock \emph{arXiv preprint arXiv:2401.07950}.

\bibitem[{Zhang et~al.(2024{\natexlab{b}})Zhang, Zhoubian, Yue, Dong, and Tang}]{zhang2024rest}
Dan Zhang, Sining Zhoubian, Yisong Yue, Yuxiao Dong, and Jie Tang. 2024{\natexlab{b}}.
\newblock Rest-mcts*: Llm self-training via process reward guided tree search.
\newblock \emph{arXiv preprint arXiv:2406.03816}.

\bibitem[{Zhang et~al.(2024{\natexlab{c}})Zhang, Zhao, Liu, Zheng, Qi, Gu, Zhang, Dong, and Tang}]{zhang2024naturalcodebench}
Shudan Zhang, Hanlin Zhao, Xiao Liu, Qinkai Zheng, Zehan Qi, Xiaotao Gu, Xiaohan Zhang, Yuxiao Dong, and Jie Tang. 2024{\natexlab{c}}.
\newblock Naturalcodebench: Examining coding performance mismatch on humaneval and natural user prompts.
\newblock \emph{arXiv preprint arXiv:2405.04520}.

\bibitem[{Zhang et~al.(2024{\natexlab{d}})Zhang, Shen, Lu, and Zhuang}]{zhang2024data}
Wenqi Zhang, Yongliang Shen, Weiming Lu, and Yueting Zhuang. 2024{\natexlab{d}}.
\newblock Data-copilot: Bridging billions of data and humans with autonomous workflow.
\newblock In \emph{ICLR 2024 Workshop on Large Language Model (LLM) Agents}.

\bibitem[{Zheng et~al.(2023)Zheng, Xia, Zou, Dong, Wang, Xue, Wang, Shen, Wang, Li et~al.}]{zheng2023codegeex}
Qinkai Zheng, Xiao Xia, Xu~Zou, Yuxiao Dong, Shan Wang, Yufei Xue, Zihan Wang, Lei Shen, Andi Wang, Yang Li, et~al. 2023.
\newblock Codegeex: A pre-trained model for code generation with multilingual evaluations on humaneval-x.
\newblock \emph{arXiv preprint arXiv:2303.17568}.

\bibitem[{Zhuo et~al.(2024)Zhuo, Vu, Chim, Hu, Yu, Widyasari, Yusuf, Zhan, He, Paul et~al.}]{zhuo2024bigcodebench}
Terry~Yue Zhuo, Minh~Chien Vu, Jenny Chim, Han Hu, Wenhao Yu, Ratnadira Widyasari, Imam Nur~Bani Yusuf, Haolan Zhan, Junda He, Indraneil Paul, et~al. 2024.
\newblock Bigcodebench: Benchmarking code generation with diverse function calls and complex instructions.
\newblock \emph{arXiv preprint arXiv:2406.15877}.

\end{thebibliography}
\bibliographystyle{acl_natbib}

\clearpage\newpage
\appendix

\section{Appendix}
\label{sec: appendix}

\subsection{Task Definition}
\label{ssec: task_des}
We define six typical data science tasks as follows:

1) \textbf{Data cleaning and preprocessing.} This task detects and processes missing values, outliers, and duplicate data; and standardizes data formats, such as a uniform format for dates and times.
    
2) \textbf{Data exploration and statistics understanding.} This task calculates basic statistical indicators of data (mean, median, standard deviation, etc.), generates data distribution charts (histograms, box plots, etc.), calculates correlations between variables, and draws correlation matrices or maps. 
    
3) \textbf{Data visualization.} The goal of this task is to visualize and analyze data and create interactive charts so users can freely explore the data.
    
4) \textbf{Predictive modeling.} The task aims to select the appropriate machine learning algorithm, such as linear regression, decision tree, random forest, etc.; carry out feature engineering, such as feature selection, feature transformation, feature combination, etc.; the data set is divided into the training set and test set, and the model is trained and evaluated; and select the corresponding evaluation indicators for different prediction problems, such as classification, regression or clustering. 
    
5) \textbf{Data mining and Pattern recognition.} This task uses association rule mining, frequent item set mining, and other methods to find interesting patterns in the data; Text mining technology is used to extract keywords, topics, and other information from text data; and apply cluster analysis, classification algorithms, etc. to identify underlying patterns and structures. 
Pattern recognition tasks can conduct these functions: image recognition, text clustering, and time series detection.
    
6) \textbf{Interpretability and Report generation}. This task aims to provide explanations of model results, such as feature importance, model parameters, etc., and automatically generate reports and summaries that present the results of the analysis in a way that is easy to understand and share.

\subsection{Comparison with Existing Benchmarks}
\label{app_ssec: comparison}
We perform correlation analysis to evaluate the alignment between \benchmark\space and coding evaluations like BCB and LCB.
To achieve correlation analysis, we calculate both Pearson’s $r$ and Spearman’s $p$ correlation coefficients, which provide insights into the strength and direction of relationships between our benchmark and these established metrics. 
This analysis not only validates our results but also ensures robustness across different evaluation dimensions. Our findings indicate strong positive correlations, suggesting that our benchmark aligns well with these established coding evaluation metrics.

\begin{table}[t!]
    \centering
    \caption{Correlation with LCB and BCB.}
    \vspace{-0.3cm}
    \label{tab: correlation}
    \resizebox{0.8\linewidth}{!}{
    \begin{tabular}{lcc} 
    \specialrule{.16em}{0pt}{.65ex}
    & \multicolumn{2}{c}{\benchmark} \\
    \cmidrule(lr){2-3} 
    \cmidrule(lr){2-3} & $r$ & $p$ \\ 
    \specialrule{.10em}{.4ex}{.65ex}
    LiveCodeBench (LCB) & 0.853 & 0.673  \\ 
    BigCodeBench (BCB) & 0.823 & 0.808 \\
    \specialrule{.16em}{.4ex}{0pt}
    \end{tabular}
    }
\end{table}

While \benchmark\space does show a correlation with LCB or BCB in Table~\ref{tab: correlation}, our benchmark offers several unique and important contributions:

\begin{itemize}[leftmargin=*,itemsep=0pt,parsep=0.5em,topsep=0.3em,partopsep=0.3em]
    \item \textbf{Domain-Specific Focus:} \benchmark\space specifically targets data science and analytics tasks. However, existing benchmarks primarily focus on general programming problems. This specialization helps evaluate models' capabilities in handling real-world data analysis scenarios.

    \item \textbf{Task Diversity:} Our \benchmark\space includes unique task types like data preprocessing, visualization, and statistical analysis. These tasks are underrepresented in current benchmarks. This provides deeper insights into models' data science-specific capabilities.

    \item \textbf{Complementary Insights:} While overall correlations exist, we observe meaningful differences in model rankings. For example, models like Meta-Llama-3-8B-Instruct and CodeLlama-34B-Instruct show distinct performance patterns. These differences highlight capabilities specific to data science tasks that other benchmarks may not capture.
\end{itemize}
\textcolor{black}{The correlation with existing benchmarks validates our evaluation methodology, while our domain-specific focus provides valuable new insights for assessing AI models in data science applications.}

\begin{figure}[t!]
    \centering
    \includegraphics[width=0.8\linewidth]{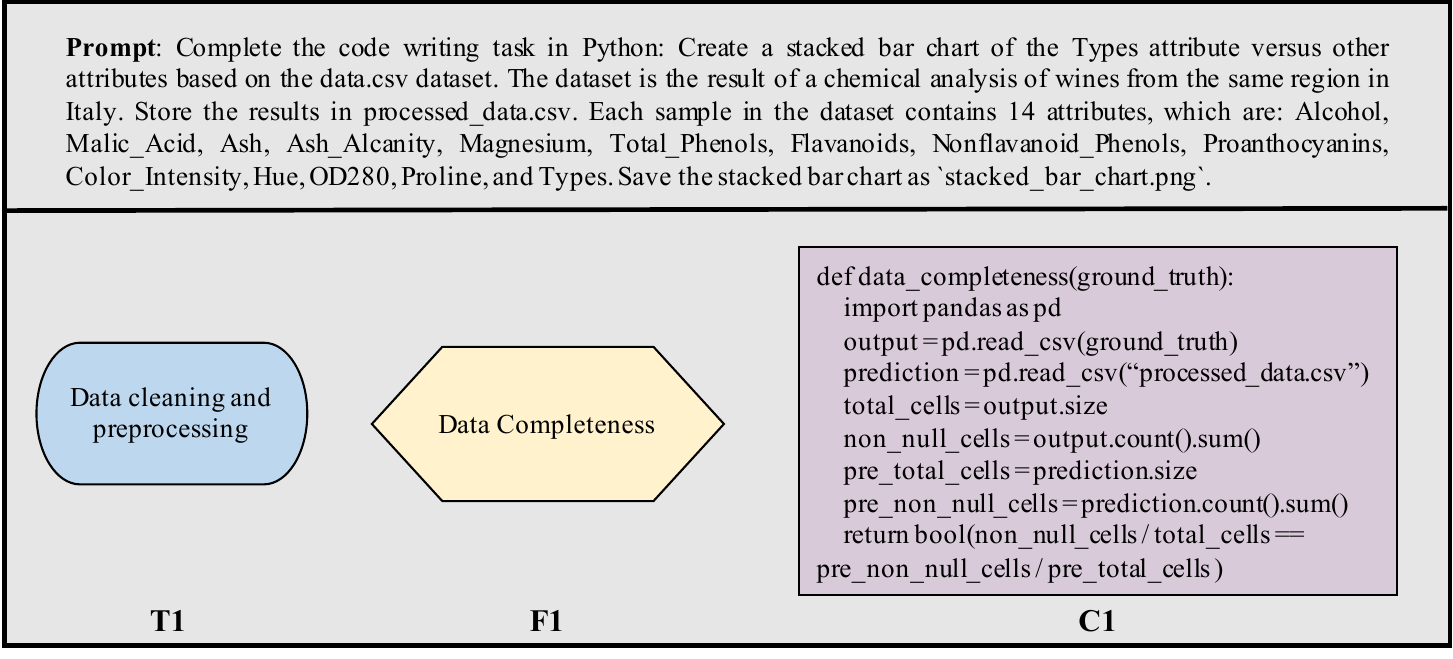}
    \caption{An example of TFC tuples.}
    \label{fig: tfc}
\end{figure}

\subsection{Motivation and Example of Task-Function-Code (TFC)}
\textcolor{black}{
The TFC framework was developed to address several critical challenges in automated evaluation of data science tasks:}
\begin{itemize}[leftmargin=*,itemsep=0pt,parsep=0.5em,topsep=0.3em,partopsep=0.3em]
    \item \textbf{Systematic Task Selection:} TFC provides a structured approach to identify and categorize key tasks across six established types. This systematic organization ensures comprehensive coverage of essential data science operations and helps maintain evaluation consistency and completeness.
    \item \textbf{Standardized Evaluation Metrics:} Data science tasks often lack standardized evaluation criteria. TFC addresses this by explicitly defining appropriate evaluation functions for each task. For example, data preprocessing tasks require specific metrics that differ from visualization tasks. This standardization ensures fair and consistent assessment.
    \item \textbf{Automated Execution Framework:} TFC includes executable code components for both tasks and evaluation metrics. This automation significantly improves evaluation efficiency, result reproducibility, and testing scalability.
    \item \textbf{Ground Truth Generation:} TFC serves as a crucial foundation for establishing ground truth, particularly valuable for complex tasks where ground truth is not readily available, and enables systematic verification and validation of model outputs.
\end{itemize}
Overall, the TFC structure in Figure~\ref{fig: tfc} represents a novel contribution by providing a comprehensive framework that bridges the gap between task definition, evaluation criteria, and automated assessment in data science contexts.

\subsection{Programmatic Rules}
We list all programmatic rules in Table~\ref{tab: program_rules}.

\begin{table*}[t!]
\centering
\caption{Details of programmatic rules.}
\vspace{-0.2cm}
\resizebox{0.97\textwidth}{!}{
\begin{tabular}{c|c|c|c|c|c|c}
\specialrule{.16em}{0pt}{.65ex}
Aggregate Function                  & Task                                    &  Type  & Rule & 
Comparison            & GT & Threshold                   \\
\specialrule{.10em}{.4ex}{.65ex}
Mean Squared Error          & Calculate MSE                                   & Decimal & Bool & It $\leq$ to GT is 1, it $>$ GT is 0 & Yes     & -                     \\
Data Cleaning Completeness  & row/column number                               & Integer & Bool & If it ==  GT, it is 1; if it != GT, it is 0  & Yes     & -                     \\
Silhouette Score            & Calculate                                & Decimal & Bool & If it $\geq$ to GT is 1, if it $<$ GT is 0 & Yes     & 0.5                  \\
Model Accuracy              & Calculate F1                                    & Decimal & Bool & If it $\geq$ to GT is 1, if it $<$ GT is 0  & Yes     & 0.9                  \\
Model Accuracy              & Calculate Accuracy                              & Decimal & Bool & If it $\geq$ to GT is 1, if it $<$ GT is 0  & Yes     & 0.9                  \\
R-squared Value             & Calculate                              & Decimal & Bool & If it $\geq$ to GT is 1, if it $<$ GT is 0  & Yes     & 0.9                  \\
Data Cleaning Completeness  & number of nulls                            & Integer & Bool & If it ==  0, returns 1; if it $>$ 0, returns 0  & Yes     & -                     \\
Data Quality Score         & number of outlier                         & Integer & Bool & If it ==  GT, it is 1; if it != GT, it is 0  & Yes     & -                     \\
Data Accuracy               & MSE                                     & Decimal & Bool & Threshold judgment          &  Yes     & 0.05                 \\
Data Completeness           & Null number ratio                                & Decimal & Bool & If it ==  GT, it is 1; if it != GT, it is 0  &  Yes     & -                      \\
Data Quality Score          & Calculate loss                                  & Decimal & Bool & It $\leq$ to GT is 1, it $>$ GT is 0 &   Yes    & -                     \\
Association Rule Confidence & Association rule accuracy                                 & Decimal & Bool & Threshold judgment          & Yes      & 0.9                  \\                \specialrule{.16em}{.4ex}{0pt}
\end{tabular}
}
\label{tab: program_rules}
\end{table*}

\subsection{Caveats when Using LLMs for Data Science}
Here we list the issues that occurred during testcase generation, most of which have been addressed by modifying the prompts. We notice that some of the issues may be disruptive to the system and some may be subtle but important.
\begin{enumerate}
    \item Be careful when using LLMs on well-known open source datasets, especially with customized tasks and data split. LLMs may memorize some open-source datasets. For example, if we want to use part of the penguin dataset that does not contain certain columns, the model (GPT-4o) will still explicitly process those columns in the code.
    \item Hallucination during data pre-processing. For example, when the model is required to merge two CSV, it may hallucinate on a common column and not go through all the columns in the files to find the actual ones.
    \item On multilingual tasks. LLMs may not be able to select the correct encoding. For instance, when they are required to open a CSV file that has content in Chinese, they will struggle to choose the correct encoding to open the file. Even if they are hinted that the file is in Chinese, they may choose encodings other than ``gbk'', e.g., ``latin''.
\end{enumerate}

\subsection{Related Works}
\label{sec: related_works}
\subsubsection{LLMs for Data Science} 
With the popularity of large-scale language models, researchers have developed a series of LLM-based agents~\cite{yao2022react, wang2024battleagentbench, xu2024androidlab, xia2024scenegenagent}. 
For data science, SheetCopilot~\citep{li2024sheetcopilot} designs a tabular agent, which directly processes natural language-described tasks, and generates and executes a series of operation plans on datasheets to produce the desired results. 
Data Copilot~\citep{zhang2024data} is an intelligent agent that serves as a bridge between users and data, which automatically executes data processing, prediction, and visualization tasks based on users' data needs.
InsightPilot~\citep{ma2023insightpilot} focuses on exploratory data analysis and can automatically discover data insights related to fuzzy questions raised by users.
Data interpreter~\citep{hong2024datainterpreter} augments problem-solving in data science with dynamic planning with hierarchical graph structures, tool integration, and logical inconsistency identification in feedback.
Furthermore, ReAct~\cite{yao2022react} is quite similar to our backbone agentic framework, DataInterpreter. Both frameworks offer foundational pipelines rather than specific evaluation metrics tailored for diverse data science tasks.
However, the correctness of data analysis in data science has a significant impact on decision-making.
Therefore, with the continuous increase of data science agents, it is urgent to conduct a comprehensive and in-depth evaluation of data science agents.

\subsubsection{LLM Agent Evaluation Benchmarks for Data Science}

Assessing the effectiveness of LLMs in handling diverse and challenging data science prompts is essential to push the boundaries of benchmarking practices in the field of data science and LLM research.
Data science agents often solve problems by generating code, so the capabilities of data science agents are closely related to the code generation capabilities of large models. 
There are already many benchmarks for evaluating the code capability of large models.
MLAgentBench~\citep{huang2023mlbenchmarking} benchmarks the LLMs' abilities on traditional machine learning tasks.
NaturalCodeBench~\citep{zhang2024naturalcodebench} evaluates the capabilities of code generation models on the real prompts from the CodeGeeX~\citep{zheng2023codegeex} platform.
However, the general code evaluation benchmark ignores the characteristics of data science tasks and cannot comprehensively and effectively evaluate the capabilities of large models in data science. 

Recently, some evaluation benchmarks for large language models in data science have been proposed. 
Text2Analysis~\citep{he2024text2analysis} constructs the evaluation benchmark to evaluate the model's ability to handle data analysis functions and fuzzy questions on tabular data. Their prompts are obtained through manual annotation and large model generation.
Furthermore, DAEval~\citep{huinfiagent} is developed as another evaluation benchmark and it contains 257 data analysis questions on CSV data and questions, which are generated by LLMs.
However, the prompts in these two works often only involve one task, and these prompts involve relatively simple data analysis operations.
In practical data science analysis tasks, user questions often involve multiple tasks and involve performing complex data analysis operations.
Therefore, we aim to provide a data science evaluation benchmark that is more in line with practical scenarios, especially for problems involving multiple subtasks and complex data analysis operations.

\subsection{Related Work in Data Science}
Recently, some evaluation benchmarks for large language models in data science have been proposed. 
Text2Analysis~\citep{he2024text2analysis} constructs the evaluation benchmark to evaluate the model's ability to handle data analysis functions and fuzzy questions on tabular data. Their prompts are obtained through manual annotation and large model generation.
Furthermore, DAEval~\citep{huinfiagent} is developed as another evaluation benchmark and it contains 257 data analysis questions on CSV data and questions, which are generated by LLMs.
However, the prompts in these two works often only involve one task, and these prompts involve relatively simple data analysis operations.
In practical data science analysis tasks, user questions often involve multiple tasks and involve performing complex data analysis operations.
Therefore, we aim to provide a data science evaluation benchmark that is more in line with practical scenarios, especially for problems involving multiple subtasks and complex data analysis operations.

\subsection{Results}
We present the detailed results for our collected prompt and BigCodeBench source in Table~\ref{tab:our_results} and Table~\ref{tab:bcb_results}.

\begin{table*}[t!]
    \centering
    \caption{Evaluation results for \benchmark$\space$ (our collected prompts).}
    \vspace{-0.2cm}
    \label{tab:our_results}
    \resizebox{0.97\textwidth}{!}{
    \begin{tabular}{c|c|c|c|c|c|c|c|c|c|c}
    \specialrule{.16em}{0pt}{.65ex}
        \multirow{2}{*}{Models} & \multirow{2}{*}{Size} & \multicolumn{2}{c|}{Coarse-grained Metrics} & \multicolumn{6}{c|}{Fine-grained Metrics} & \multirow{2}{*}{Score}
        \\
        \cmidrule{3-10}
        & & Success (\%) & CR (\%) & VLM & F1 & F2 & F3 & F4 & F5 & \\
        \specialrule{.10em}{.4ex}{.65ex}
        o1-mini                      & N/A                      & 13.45  & 15.43      & 2.87 & 53.75 & \text{0.00}  & 45.50 & 46.25 & 25.79 & 19.41 \\
        GPT-4o-2024-05-13            & N/A                      & \textbf{19.82}  & \textbf{17.89}      & \textbf{3.91} & \textbf{71.79} & \text{0.00}  & \textbf{54.50} & \underline{70.00} & 33.95 & \textbf{24.32} \\
        GPT-4o-mini                  & N/A                      & 12.73  & 17.35      & 3.05 & 62.86 & \textbf{15.00} & \underline{50.50} & 69.38 & \textbf{43.68} & 24.13 \\
        GPT-4-Turbo                  & N/A                      & \underline{17.27}  & \underline{17.36}      & \underline{3.09} & \underline{65.18} & 10.00 & 44.50 & \textbf{78.12} & \underline{41.05} & \underline{24.24} \\
        Claude-3-5-Sonnet-20240620   & N/A                      & 8.00      & 11.12      & 2.14 & 32.32 & \text{0.00}  & 36.00 & 63.12 & 17.63 & 15.19 \\
        GLM-4-Flash                  & N/A                      & 9.82   & 7.43       & 1.33 & 47.14 & \text{0.00}  & 22.00 & 24.38 & 3.95  & 10.26 \\
        \specialrule{.10em}{.4ex}{.65ex}
        Meta-Llama-3.1-8B-Instruct   & 8B                       & 10.00     & 7.72       & 1.29 & 45.18 & \text{0.00}  & 16.00 & 23.12 & 8.95  & 10.24 \\
        Meta-Llama-3-8B-Instruct     & 8B                       & 1.64   & 1.43       & 0.52 & 7.86  & \text{0.00}  & 4.50  & 6.88  & \text{0.00}  & 2.00  \\
        Gemma-2-9B-it                & 9B                       & 5.64   & 5.51       & 1.06 & 26.79 & \text{0.00}  & 13.00 & 22.50 & 2.89  & 7.18  \\
        GLM-4-9B-Chat                & 9B                       & 10.55  & 9.96       & 1.69 & 55.36 & \text{0.00}  & 31.00 & 28.75 & 21.32 & 13.91 \\
        Qwen2.5-7B-Instruct          & 7B                       & 11.64  & 10.11      & 1.43 & 55.36 & \text{0.00}  & 36.50 & 33.12 & 18.42 & 14.40 \\
        Qwen2-7B-Instruct            & 7B                       & 6.91   & 5.90       & 1.16 & 32.50 & \text{0.00}  & 18.00 & 21.88 & 2.37  & 7.97  \\
        Qwen2-1.5B-Instruct          & 1.5B                     & 1.82   & 1.60       & 0.40 & 3.57  & \text{0.00}  & 2.00  & 13.12 & 0.79  & 2.13  \\
        Yi-1.5-9B-Chat-16K           & 9B                       & 6.18   & 4.25       & 0.73 & 30.36 & \text{0.00}  & 16.00 & 8.75  & 3.95  & 6.06  \\
        \specialrule{.10em}{.4ex}{.65ex}
        CodeLlama-34B-Instruct    & 34B                      & \text{0.00}      & 0.03       & \text{0.00} & \text{0.00}  & \text{0.00}  & \text{0.00}  & \text{0.00}  & \text{0.00}  & 0.02  \\
        CodeLlama-13B-Instruct    & 13B                      & 0.73   & 0.50       & 0.04 & 4.46  & \text{0.00}  & \text{0.00}  & 3.75  & \text{0.00}  & 0.77  \\
        CodeLlama-7B-Instruct     & 7B                       & 0.55   & 0.27       & \text{0.00} & 1.96  & \text{0.00}  & \text{0.00}  & \text{0.00}  & \text{0.00}  & 0.30  \\
        StarCoder2-15B               & 15B                      & 0.18   & 0.20       & 0.07 & 0.54  & \text{0.00}  & \text{0.00}  & 0.62  & \text{0.00}  & 0.20 \\
        Deepseek-Coder-33B-instruct  & 33B                      & 12.55  & 13.53      & 2.29 & 62.86 & \text{0.00}  & 43.00 & 51.88 & 21.32 & 18.49 \\
        Deepseek-Coder-6.7B-instruct & 6.7B                     & 12.55  & 13.56      & 1.93 & 63.21 & \text{0.00}  & 39.00 & 53.75 & 21.05 & 18.39 \\
        Deepseek-Coder-1.3B-instruct & 1.3B                     & 0.73   & 0.61       & 0.10 & 3.39  & \text{0.00}  & \text{0.00}  & 1.25  & \text{0.00}  & 0.67  \\
        Qwen2.5-Coder-7B-Instruct    & 7B                       & 6.18   & 7.87       & 1.48 & 40.18 & \text{0.00}  & 27.50 & 33.75 & 4.47  & 10.79 \\
        Qwen2.5-Coder-1.5B-Instruct  & 1.5B                     & 6.18   & 7.52       & 0.81 & 38.57 & \text{0.00}  & 15.50 & 40.00 & 10.53 & 10.46 \\
        \specialrule{.16em}{.4ex}{0pt}
    \end{tabular}
    }
\end{table*}

\begin{table*}[t!]
    \centering
    \caption{Evaluation results for \benchmark$\space$ (BigCodeBench source).}
    \vspace{-0.2cm}
    \label{tab:bcb_results}
    \resizebox{0.97\textwidth}{!}{
    \begin{tabular}{c|c|c|c|c|c|c|c|c|c}
    \specialrule{.16em}{0pt}{.65ex}
    \multirow{2}{*}{Models} & \multirow{2}{*}{Size} & \multicolumn{2}{c|}{Coarse-grained Metrics} & \multicolumn{5}{c|}{Fine-grained Metrics} & \multirow{2}{*}{Score} \\
    \cmidrule{3-9}
    & & Success (\%) & CR (\%) & F1 & F2 & F3 & F4 & F5 & \\
    \specialrule{.10em}{.4ex}{.65ex}
    o1-mini                      & N/A                      & 35.15  & 55.08      & 41.62 & 25.62 & 32.89 & 25.90 & 23.16 & 47.77 \\
    GPT-4o-2024-05-13            & N/A                      & \textbf{81.62}  & \textbf{85.09}      & \textbf{77.30} & \textbf{74.63} & \textbf{74.21} & \textbf{71.79} & \textbf{65.48} & \textbf{81.81} \\
    GPT-4o-mini                  & N/A                      & 63.11  & 71.10       & 59.46 & 58.89 & 60.26 & 55.90 & 56.79 & 67.49 \\
    GPT-4-Turbo                  & N/A                      & 63.35  & 72.54      & 61.35 & 52.04 & 62.11 & 54.36 & 53.95 & 68.14 \\
    Claude-3-5-Sonnet-20240620   & N/A                      & 60.48  & 73.59      & 54.59 & 49.11 & 62.37 & 49.49 & 55.39 & 68.08 \\
    GLM-4-Flash                  & N/A                      & 37.07  & 42.8       & 33.04 & 39.11 & 36.05 & 28.72 & 17.89 & 39.55 \\
    \specialrule{.10em}{.4ex}{.65ex}
    Meta-Llama-3.1-8B-Instruct   & 8B                       & 29.58  & 42.51      & 35.95 & 24.26 & 23.95 & 22.82 & 31.41 & 38.16 \\
    Meta-Llama-3-8B-Instruct     & 8B                       & 3.29   & 4.74       & 2.97  & 1.67  & 2.11  & 1.28  & 1.96  & 3.98  \\
    Gemma-2-9B-it                & 9B                       & 7.54   & 12.81      & 25.95 & 22.46 & 27.37 & 16.67 & 21.84 & 15.06 \\
    GLM-4-9B-Chat                & 9B                       & 30.72  & 37.11      & 23.65 & 30.78 & 27.11 & 26.67 & 18.42 & 33.84 \\
    Qwen2.5-7B-Instruct          & 7B                       & 54.43  & 64.12      & 49.80 & 48.40 & 50.79 & 49.23 & 40.15 & 59.52 \\
    Qwen2-7B-Instruct            & 7B                       & 28.08  & 32.06      & 30.41 & 27.63 & 32.26 & 27.18 & 9.21  & 30.18 \\
    Qwen2-1.5B-Instruct          & 1.5B                     & 4.67   & 6.73       & 4.86  & 2.63  & 3.68  & 3.33  & 6.00  & 5.97  \\
    Yi-1.5-9B-Chat-16K           & 9B                       & 48.74  & 54.9       & 40.70 & 48.34 & 42.11 & 46.41 & 35.64 & 51.53 \\
    \specialrule{.10em}{.4ex}{.65ex}
    CodeLlama-34B-Instruct    & 34B                      & 1.20    & 1.94       & 1.35  & 1.11  & 2.63  & 2.05  & 1.58  & 1.85  \\
    CodeLlama-13B-Instruct    & 13B                      & 13.71  & 19.3       & 14.05 & 15.07 & 12.53 & 17.95 & 6.84  & 17.52 \\
    CodeLlama-7B-Instruct     & 7B                       & 3.65   & 5.19       & 4.05  & 3.15  & 3.42  & 2.31  & 2.11  & 4.57  \\
    StarCoder2-15B               & 15B                      & 2.69   & 3.41       & 3.24  & 2.41  & 2.11  & 4.36  & 1.58  & 3.21  \\
    Deepseek-Coder-33B-instruct  & 33B                      & \underline{70.12}  & \underline{76.94}      & \underline{66.58} & \underline{62.63} & \underline{63.16} & \underline{64.87} & \underline{57.59} & \underline{73.11} \\
    Deepseek-Coder-6.7B-instruct & 6.7B                     & 45.09  & 50.86      & 37.00 & 45.96 & 48.79 & 44.10 & 17.11 & 47.50 \\
    Deepseek-Coder-1.3B-instruct & 1.3B                     & 20.36  & 25.05      & 16.22 & 19.44 & 17.63 & 21.28 & 10.53 & 22.81 \\
    Qwen2.5-Coder-7B-Instruct    & 7B                       & 58.02  & 68.01      & 55.34 & 57.44 & 49.26 & 45.38 & 45.36 & 63.15 \\
    Qwen2.5-Coder-1.5B-Instruct  & 1.5B                     & 28.20   & 35.60       & 26.94 & 28.96 & 26.74 & 26.15 & 18.32 & 32.69 \\
    \specialrule{.16em}{.4ex}{0pt}
    \end{tabular}
    }
\end{table*}

\subsection{AI Assistants In Writing}
While writing this paper, we employed AI to rectify grammatical errors and revise unreasonable descriptions.

\subsection{VLM-as-a-judge Prompt and Examples}
\label{sec:vlm_prompt}
In this section, we define the criteria to measure the quality of LLM-generated figures/charts.

\onecolumn
\begin{tcolorbox}[colback=blue!5!white, colframe=blue!75!black, title=Prompt for VLM-as-a-judge, text width=\textwidth]
Above are two figures, which are A and B. The first figure is the ground truth image and the second figure is the predicted image. The total score is 5. Please score B following the criteria below:\\
- add 1 point for Data Representation Consistency: Ensure that the underlying data represented by the two charts is identical. This includes the values for all data points and the range of the data. Any variation in the dataset used would make the charts different.\\
- add 1 point for Axis Labels and Scales: Verify that both charts have identical axis labels, units, and scales. Any difference in how the axes are labeled or scaled, such as using logarithmic vs. linear scales, can affect the interpretation of the data.\\
- add 1 point for Graphical Elements: Check if the visual elements (such as lines, bars, markers, etc.) are represented the same way in both charts. Line thickness, marker styles, and colors should match across charts for them to be considered visually equal.\\
- add 1 point for Legend and Annotations: Confirm that any legends, titles, or annotations (e.g., text labels, arrows, or highlights) are the same in both charts. These elements often provide crucial context for interpreting the chart.\\
- add 1 point for Chart Dimensions and Layout: Ensure that the dimensions (height and width), aspect ratios, and layout of the charts are identical. Even if the content and representation are similar, a different aspect ratio or spacing between elements can change the chart’s overall appearance and interpretation.\\
\\
Please write down the total score for B based on the criteria above, and provide a brief explanation of your reasoning. If you believe that the two figures are not identical, please explain the differences you observed.\\
\\
\#\#\# Explanation:\\
your explanation here\\
\\
\#\#\# Total Score:\\
x/5
\end{tcolorbox}
\clearpage
\onecolumn
\begin{tcolorbox}[colback=blue!5!white, colframe=blue!75!black, title=Example \#1 of VLM-as-a-judge (Model: Claude-3-5-sonnet-20240620), text width=\textwidth]
Image A:\\
\begin{center}
    \includegraphics[width=0.5\textwidth]{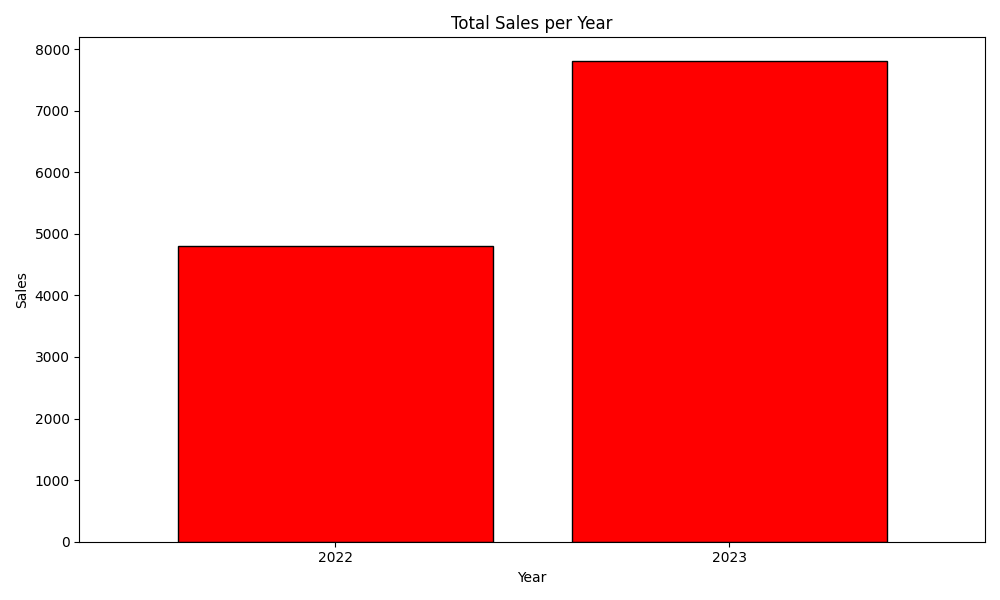}
\end{center}
Image B:\\
\begin{center}
    \includegraphics[width=0.5\textwidth]{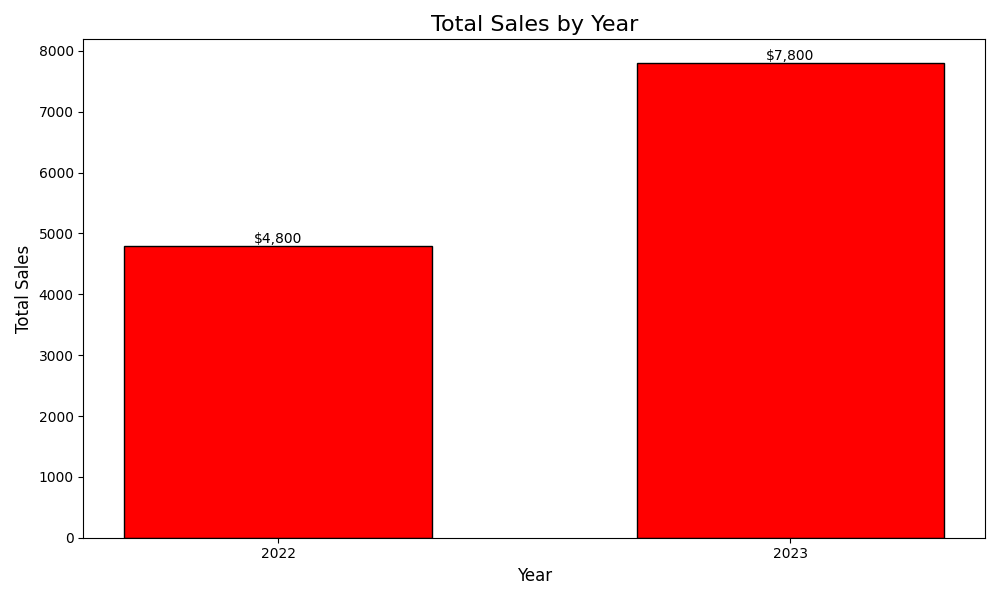}
\end{center}
[Prompt for VLM-as-a-judge]\\
\#\#\# Explanation:\\
1. **Data Representation Consistency**: Both charts represent the same underlying data for the years 2022 and 2023, with sales figures of \$4,800 for 2022 and \$7,800 for 2023. **+1 point**.\\
\\
2. **Axis Labels and Scales**: Both charts have identical X-axis labels (years) and Y-axis labels (total sales). They both use the same scale on the Y-axis, ranging from 0 to 8000. **+1 point**.\\
\\
3. **Graphical Elements**: The bars in both charts are red with the same thickness. Thus, the visual representation matches. **+1 point**.\\
\\
4. **Legend and Annotations**: The second chart includes dollar amounts on each bar, which isn’t present in the first chart. This difference impacts the context provided for interpreting the chart, reducing consistency. **0 points**.\\
\\
5. **Chart Dimensions and Layout**: Both charts appear similar in terms of layout and dimensions, with no significant visual differences observed. **+1 point**.\\
\\
Given the above evaluations, the total score for B is calculated as follows:\\
\\
\#\#\# Total Score:\\
4/5
\end{tcolorbox}
\twocolumn

\onecolumn
\begin{tcolorbox}[colback=blue!5!white, colframe=blue!75!black, title=Example \#2 of VLM-as-a-judge (Model: CodeLlama-13B-Instruct), text width=\textwidth]
Image A:\\
\begin{center}
    \includegraphics[width=0.5\textwidth]{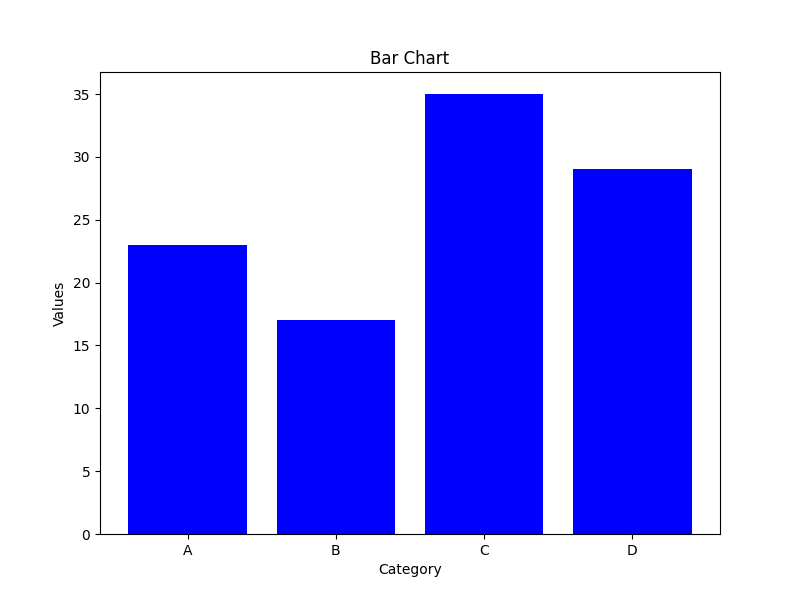}
\end{center}
Image B:\\
\begin{center}
    \includegraphics[width=0.5\textwidth]{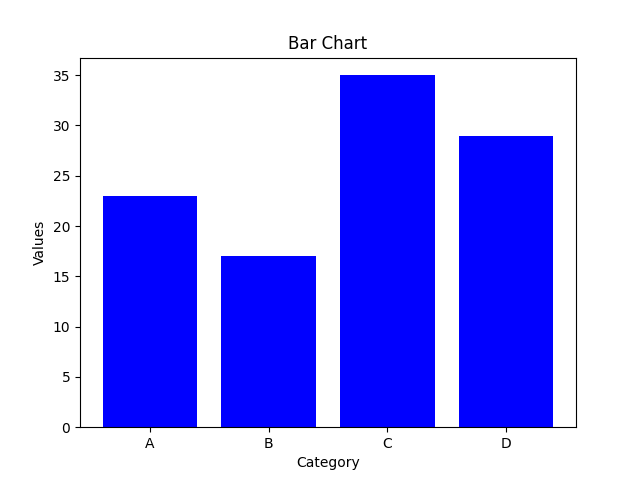}
\end{center}
[Prompt for VLM-as-a-judge]\\
\#\#\# Explanation:\\
1. **Data Representation Consistency**: Both figures (A and B) represent the same underlying data with identical values for categories A, B, C, and D. Therefore, 1 point is awarded.\\
2. **Axis Labels and Scales**: The axis labels ("Category" for the x-axis and "Values" for the y-axis) appear to be identical in both images, and the scale for the y-axis seems consistent. Thus, 1 point is awarded.\\
3. **Graphical Elements**: Both charts use blue bars to represent the data, and the thickness and style of the bars are similar. Therefore, 1 point is awarded.\\
4. **Legend and Annotations**: There are no legends or additional annotations in either image, and the titles are the same ("Bar Chart"). Thus, 1 point is awarded.\\
5. **Chart Dimensions and Layout**: The dimensions and overall layout of the two charts differ. Image A has a different aspect ratio compared to Image B, which affects the total layout. Hence, 0 points are awarded.\\
\\
\#\#\# Total Score:\\
4/5
\end{tcolorbox}
\twocolumn

\onecolumn
\begin{tcolorbox}[colback=blue!5!white, colframe=blue!75!black, title=Example \#3 of VLM-as-a-judge (Model: o1-mini), width=\textwidth]
\begin{minipage}[t]{\columnwidth}
\textbf{Image A:}
\begin{center}
    \includegraphics[width=0.6\columnwidth]{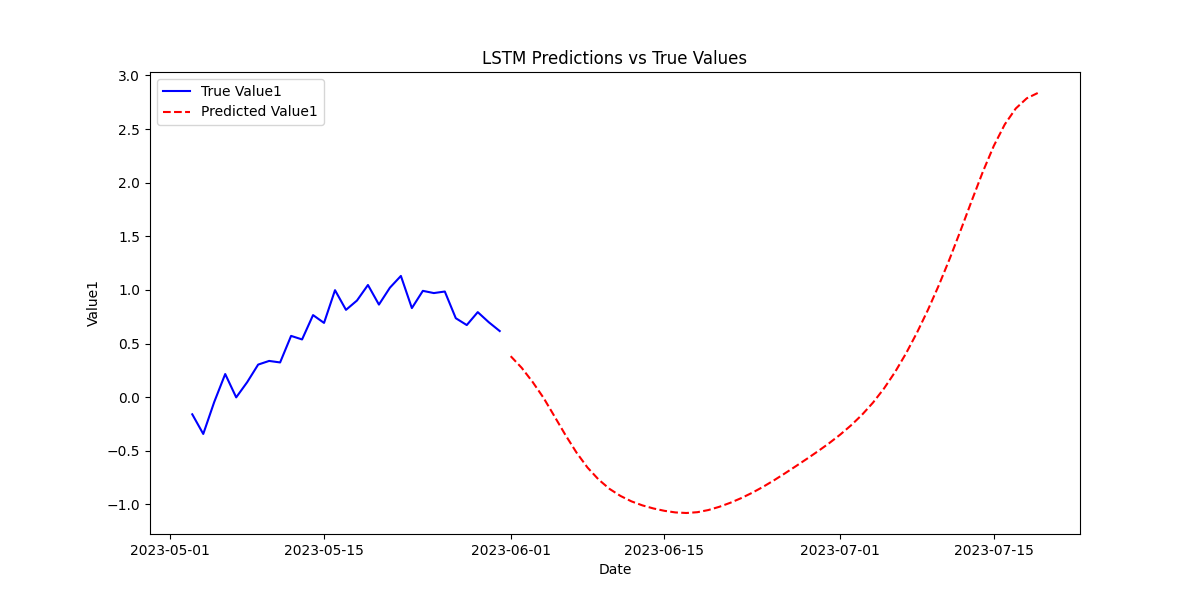}
\end{center}
\textbf{Explanation:}
\begin{enumerate}[leftmargin=*]
    \item \textbf{Data Representation Consistency:}  
    Image A represents true values compared to predictions over a broader date range, while Image B only shows predicted values for a specific date range (2023-06-01 to 2023-07-19). The data sets are not identical. \textbf{(Score: 0)}
    
    \item \textbf{Axis Labels and Scales:}  
    Both images have similar axis labels ("Date" for x-axis, "Value1" for y-axis), but Image B lacks clarity on the date range. Image A provides a wider range, with different intervals. Scales are inconsistent. \textbf{(Score: 0)}
\end{enumerate}
\end{minipage}

\begin{minipage}[t]{\columnwidth}
\textbf{Image B:}
\begin{center}
    \includegraphics[width=0.6\columnwidth]{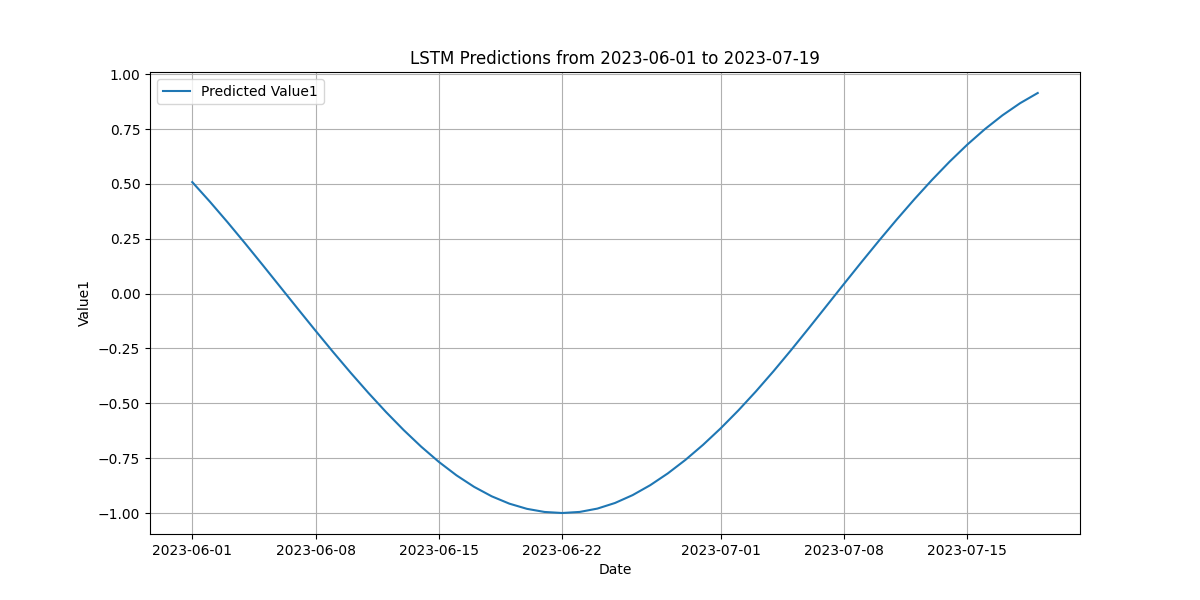}
\end{center}
\textbf{Explanation (cont.):}
\begin{enumerate}[leftmargin=*]
    \setcounter{enumi}{2} 
    \item \textbf{Graphical Elements:}  
    Image A uses a dashed red line for predictions, while Image B uses a solid blue line. This difference affects visual representation. \textbf{(Score: 0)}
    
    \item \textbf{Legend and Annotations:}  
    Image A includes a detailed legend, while Image B only identifies the predicted values. Titles are also different, affecting context and interpretability. \textbf{(Score: 0)}
    
    \item \textbf{Chart Dimensions and Layout:}  
    Image A has a wider aspect ratio for a broader date range, while Image B is more condensed. \textbf{(Score: 0)}
\end{enumerate}

\textbf{Total Score:} 0/5
\end{minipage}
\end{tcolorbox}
\twocolumn

\subsection{Qualified Prompts}
\begin{itemize}
    \item \textbf{Original Prompt 1:}    

There is a dataset with missing values in a CSV file, which records the region, height, weight, age, and salary of 36 individuals. Please address the following issues: \\(a) Calculate the proportion of missing values in each column and select the rows with at least two non-missing values in the last three columns. \\(b) Please fill in the weight column reasonably by combining the data from the height and region columns.
    
    \item \textbf{Qualified Prompt 1:}

\begin{tcolorbox}[colback=blue!5!white, colframe=blue!75!black, title=Qualified Prompt 1:]
Given a dataset with missing values in a file named \texttt{'data.csv'}which records the region, height, weight, age, and salary of 36 individuals, please address the following issues:\\
(a) Calculate the proportion of missing values in each column and select the rows with at least two non-missing values in the last three columns. Save your output in a CSV file named \texttt{'missing\_values\_proportion\\.csv'}.\\
(b) Fill in the weight column reasonably by combining the data from the height and region columns. Save this updated dataset in a CSV file named \texttt{'updated\_data.csv'}.
\end{tcolorbox}

\item \textbf{Original Prompt 2:}
You are required to analyze and visualize the "Global Terrorism Database" from Kaggle. Please load the dataset and perform data cleaning by handling missing values, removing duplicates, and correcting any anomalies. Conduct an exploratory data analysis (EDA) to understand the distribution and relationships within the dataset. Calculate basic statistical indicators such as mean, median, standard deviation, and provide summary statistics for key features like attack type, target type, and region. Generate visualizations to uncover patterns and insights. Create histograms and box plots to display the distribution of numerical features, and bar plots to show the frequency of categorical variables. Use scatter plots and heatmaps to visualize relationships and correlations between features. Identify patterns in the data related to terrorist activities. For instance, determine trends over time, geographical hotspots, and common attack methods. Use clustering techniques ( K-means clustering) to identify patterns and group similar incidents together.

    \item \textbf{Qualified Prompt 2:}

\begin{tcolorbox}[colback=blue!5!white, colframe=blue!75!black, title=Qualified Prompt 2:, text width=\textwidth]
You are required to analyze and visualize the \textit{Global Terrorism Database} from Kaggle. Please follow the steps below:
\begin{enumerate}
    \item Load the dataset \\
    Input: \texttt{gtd.csv} \\
    Output: \texttt{loaded\_data.csv} (This should contain the original data loaded without any modifications.)
    
    \item Data Cleaning 
    \begin{itemize}
        \item Handle missing values
        \item Remove duplicates
        \item Correct anomalies
    \end{itemize}
    Input: \texttt{loaded\_data.csv} \\
    Output: \texttt{cleaned\_data.csv} (This should reflect the cleaned dataset, ready for analysis.)
    
    \item Exploratory Data Analysis (EDA) 
    \begin{itemize}
        \item Calculate basic statistical indicators such as mean, median, and standard deviation
        \item Provide summary statistics for key features (attack type, target type, region)
    \end{itemize}
    Input: \texttt{cleaned\_data.csv} \\
    Output: \texttt{eda\_summary\_statistics.csv} (This should include all calculated statistics for key features.)
    
    \item Generate Visualizations 
    \begin{itemize}
        \item Create histograms and box plots for numerical features
        \item Generate bar plots for categorical variables
        \item Use scatter plots and heatmaps to visualize relationships and correlations
    \end{itemize}
    Input: \texttt{cleaned\_data.csv} \\
    Output: \texttt{visualizations.pdf} (This should include all visualizations generated in a single PDF file.)
    
    \item Identify Patterns in Data Related to Terrorist Activities 
    \begin{itemize}
        \item Determine trends over time
        \item Identify geographical hotspots
        \item Analyze common attack methods
    \end{itemize}
    Input: \texttt{cleaned\_data.csv} \\
    Output: \texttt{patterns\_analysis.csv} (This should summarize the identified patterns, trends, and hotspots.)
    
    \item Clustering Techniques 
    \begin{itemize}
        \item Use K-means clustering to identify patterns and group similar incidents
    \end{itemize}
    Input: \texttt{cleaned\_data.csv} \\
    Output: \texttt{clustering\_results.csv} (This should include the results of the clustering analysis, showing which group each incident belongs to.)
\end{enumerate}

Ensure that each output file reflects the quality of the completion of the respective subtask for further evaluation.
\end{tcolorbox}
    
\end{itemize}

\clearpage\newpage

\subsection{Modified Prompts}

\begin{itemize}
    \item \textbf{Original Prompt 1:}

\begin{tcolorbox}[colback=blue!5!white, colframe=blue!75!black, title=Modified Prompt 1:, text width=\textwidth]
Searches a directory for CSV files matching a given regular expression pattern, reads sales data from these files, and plots the sales data with month on the x-axis and sales on the y-axis.\\
Note that: Each CSV file contains two columns: \texttt{Month} and \texttt{Sales}.\\
The function should output with: \\
A list of \texttt{matplotlib.axes.\_axes.Axes} objects, each representing a plot of sales data from a matched CSV file.\\
You should write self-contained code starting with:
\begin{verbatim}
import os
import pandas as pd
import re
import matplotlib.pyplot as plt
def task_func(directory: str, pattern: str) -> list:
\end{verbatim}
\end{tcolorbox}

    \item \textbf{Modified Prompt 1:}

\onecolumn
\begin{tcolorbox}[colback=blue!5!white, colframe=blue!75!black, title=Modified Prompt 1:, text width=\textwidth]
Search a directory for CSV files matching a given regular expression pattern, read sales data from these files, and plot the sales data with month on the x-axis and sales on the y-axis.\\

Input Requirements:
\begin{itemize}
    \item Input Directory: \texttt{data}.
    \item Input Pattern: \texttt{"csv\_\textbackslash d+\textbackslash.csv"}.
\end{itemize}

Output Requirements:
\begin{enumerate}
    \item A list of \texttt{matplotlib.axes.\_axes.Axes} objects representing the plot of sales data from each matched CSV file.
    \item Save each plot as a separate image file:
    \begin{itemize}
        \item File format: PNG
        \item Output filenames: \texttt{"sales\_plot\_<filename>.png"} where \texttt{<filename>} is the name of the CSV file without the extension.
    \end{itemize}
\end{enumerate}
Input File Specification:
\begin{itemize}
    \item Each CSV file should contain two columns: \texttt{'Month'} and \texttt{'Sales'}. The input files will be located in the specified directory.
\end{itemize}

You should write self-contained code starting with:
\begin{verbatim}
import os
import pandas as pd
import re
import matplotlib.pyplot as plt

def task_func(directory: str, pattern: str) -> list:
\end{verbatim}

\end{tcolorbox}

\clearpage
\newpage

    \item \textbf{Original Prompt 2:}

\begin{tcolorbox}[colback=blue!5!white, colframe=blue!75!black, title=Modified Prompt 1:, text width=\textwidth]
Plot a scatter graph of tuples and highlight the tuple with the maximum value at index 1.\\
The function should output with: \\
\texttt{matplotlib.axes.Axes}: The Axes object of the plot for further manipulation and testing, with the title 'Max Tuple Highlighted', x-axis labeled 'x', y-axis labeled 'y', and a legend.\\
You should write self-contained code starting with:
\begin{verbatim}
import numpy as np
from operator import itemgetter
import matplotlib.pyplot as plt
def task_func(data):
\end{verbatim}
\end{tcolorbox}

    \item \textbf{Modified Prompt 2:}
    
\begin{tcolorbox}[colback=blue!5!white, colframe=blue!75!black, title=Modified Prompt 2:, text width=\textwidth]
Plot a scatter graph of tuples and highlight the tuple with the maximum value at index 1 using the input data from \texttt{"data.csv"}. The function should output the following: \\

A scatter plot saved as \texttt{"scatter\_plot.png"} with the title \texttt{'Max Tuple Highlighted'}, x-axis labeled \texttt{'x'}, y-axis labeled \texttt{'y'}, and a legend. The highlighted point should signify the tuple with the maximum value at index 1. \\

Please write self-contained code starting with:
\begin{verbatim}
import numpy as np
from operator import itemgetter
import matplotlib.pyplot as plt
def task_func(data):
\end{verbatim}
\end{tcolorbox}

\end{itemize}

\subsection{Prompt examples of different difficulty levels}
\begin{tcolorbox}[colback=blue!5!white, colframe=blue!75!black, title=Easy-level Prompt \#1:, text width=\textwidth]
\{ \\
    \texttt{"prompt":} \texttt{"}Read the dataset (input file: \texttt{"}Fish.csv\texttt{"}). Encode the dataset to divide it into training and test sets. From the dataset's four categories of Bream, Roach, Parkki, and Perch, randomly select 2 samples from each category for the test set. The remaining samples will be used as the training set. Output the number of samples in the training and test sets in a CSV file named \texttt{"}sample\_counts.csv\texttt{"}.\\ \\ Next, implement the KNN algorithm with K values of 1, 5, 15, and 100 to classify all samples in the test set. Output the classification results of the test samples to a CSV file named \texttt{"}classification\_results.csv\texttt{"}.\texttt{"}, 
    \\
    \\
    \texttt{"data\_source\_type"}: \texttt{"}2=open source data\texttt{"}
\\
\}
\end{tcolorbox}

\begin{tcolorbox}[colback=blue!5!white, colframe=blue!75!black, title=Easy-level Prompt \#2:, text width=\textwidth]
\{ 
\\
    \texttt{"prompt":} \texttt{"}Using the dataset of house prices and economic indicators provided below, please complete the following subtasks. Each subtask should generate an output file for evaluation, with specific input files where applicable.\\ \\ Dataset: \texttt{"}data.csv\texttt{"}\\ \\Subtasks:\\ \\ 1. Gaussian Naive Bayes Training: Train a Gaussian Naive Bayes model on the cleaned dataset and evaluate its accuracy.  \\   Input: \texttt{"}data.csv\texttt{"}  \\   Output: \texttt{"}gaussian\_model\_accuracy.csv\texttt{"} (contains accuracy metrics)\\ \\2. Multinomial Naive Bayes Training: Train a Multinomial Naive Bayes model on the cleaned dataset and evaluate its accuracy.  \\   Input: \texttt{"}data.csv\texttt{"}  \\   Output: `multinomial\_model\_accuracy.csv` (contains accuracy metrics)\\ \\ 3. Comparison of Models: Generate a comparative analysis of the accuracy of both models in a visual format.  \\   Input: \texttt{"}gaussian\_model\_accuracy.csv\texttt{"}, \texttt{"}multinomial\_model\_accuracy.csv\texttt{"}  \\   Output: \texttt{"}model\_comparison.png\texttt{"} \\ \\ Please ensure that the outputs contain necessary details for further evaluation.\texttt{"}, \\ \\
    \texttt{"data\_source\_type"}: \texttt{"}3=human written data\texttt{"}
\\
\}
\end{tcolorbox}

\begin{tcolorbox}[colback=blue!5!white, colframe=blue!75!black, title=Medium-level Prompt:, text width=\textwidth]
\{ \\
    \texttt{"prompt":} \texttt{"}Given a DataFrame \texttt{'df'} with sales data containing columns: TransactionID, ProductID, Quantity, SaleDate, StoreID, Revenue:\\\\1. Handle missing Revenue values: Impute any missing Revenue values with the median of the Revenue column.  \\   Input file: \texttt{'data.csv'}  \\   Output: \texttt{'step1\_imputed\_revenue.csv'} (contains the DataFrame after handling missing values)\\\\2. Detect and replace outliers: Detect and replace outliers in Quantity and Revenue. Outliers are defined as values below the 1st percentile or above the 99th percentile. Replace them with the corresponding 1st or 99th percentile value instead of removing them.  \\   Input file: \texttt{'step1\_imputed\_revenue.csv'}  \\   Output: \texttt{'step2\_replaced\_outliers.csv'} (contains the DataFrame after outlier replacement)\\\\3. Normalize Quantity and Revenue: Normalize Quantity and Revenue using Z-score normalization.  \\   Input file: \texttt{'step2\_replaced\_outliers.csv'}  \\   Output: \texttt{'step3\_normalized\_data.csv'} (contains the DataFrame after normalization)\\\\4. Ensure SaleDate format: Ensure SaleDate is in datetime format.  \\   Input file: \texttt{'step3\_normalized\_data.csv'}  \\   Output: \texttt{'step4\_formatted\_dates.csv'} (contains the DataFrame after ensuring datetime format)\\\\5. Encode ProductID and StoreID: Encode the ProductID and StoreID columns using separate label encoders to avoid any potential overlap in numerical values between categories from different columns.  \\   Input file: \texttt{'step4\_formatted\_dates.csv'}  \\   Output: \texttt{'final\_cleaned\_data.csv'} (contains the final cleaned DataFrame)\\\\Perform the specified data cleaning and preprocessing tasks and output the cleaned DataFrame as the final result.\texttt{"}, \\\\
    \texttt{"data\_source\_type"}: \texttt{"}3=human written data\texttt{"}
\\
\}
\end{tcolorbox}

\begin{tcolorbox}[colback=blue!5!white, colframe=blue!75!black, title=Hard-level Prompt:, text width=\textwidth]
\{ \texttt{"prompt":} \texttt{"}Write a Graph Recurrent Neural Network (GRNN) model based on attention mechanisms using Python for processing and analyzing time series data. Ensure to meet the following requirements:\\ \\1. \texttt{"}Graph network design\texttt{"}: Create a graph network where each graph represents an aerial formation, and the number of nodes corresponds to the number of vehicles in the formation.\\   - \texttt{"}Output\texttt{"}: Graph structure representation file.\\   - \texttt{"}Output filename\texttt{"}: \texttt{"}graph\_structure.json\texttt{"}\\ \\2. \texttt{"}Data format\texttt{"}: The time series data of aerial target formations is stored in Excel files located in the \texttt{"}.data/\texttt{"} directory, where each Excel file contains multiple sheets, with each sheet representing the time series data of a vehicle.\\   - \texttt{"}Input\texttt{"}: Excel files from \texttt{"}..data/\texttt{"} directory.\\   - \texttt{"}Input filename\texttt{"}: from \texttt{"}data\_1.xlsx\texttt{"} to \texttt{"}data\_5.xlsx\texttt{"}\\ \\3. \texttt{"}Data reading\texttt{"}: Read all Excel files from the \texttt{"}..data/\texttt{"} directory and extract the sheets for processing.\\   - \texttt{"}Output\texttt{"}: Combined time series data in a structured format.\\   - \texttt{"}Output filename\texttt{"}: \texttt{"}combined\_data.csv\texttt{"}\\ \\ 4. \texttt{"}Data segmentation\texttt{"}: Segment the time series data of each vehicle using a sliding window with length \texttt{"}l\texttt{"} and step size \texttt{"}s\texttt{"}.\\   - \texttt{"}Output\texttt{"}: Segmented time series data.\\   - \texttt{"}Output filename\texttt{"}: \texttt{"}segmented\_data.csv\texttt{"}\\\\5. \texttt{"}Training and testing set division\texttt{"}: Divide the segmented data into training and testing sets with a 7:3 ratio.\\   - \texttt{"}Output\texttt{"}: Training and testing datasets.\\   - \texttt{"}Output filenames\texttt{"}: \texttt{"}train\_data.csv\texttt{"}, \texttt{"}test\_data.csv\texttt{"} \\\\6. \texttt{"}Labels\texttt{"}: The last column of each Excel file contains label data, which should be preserved during the reading and segmentation process. \\\\7. \texttt{"}Model training\texttt{"}: Train the GRNN model using the training set to identify the action intentions of aerial target formations.\\   - \texttt{"}Output\texttt{"}: Trained model file.\\   - \texttt{"}Output filename\texttt{"}: \texttt{"}trained\_model.h5\texttt{"}\\\\8. \texttt{"}Training parameters\texttt{"}: Set the number of training epochs to 60, and record the loss value, accuracy, recall, and F1-Score for each epoch.\\   - \texttt{"}Output\texttt{"}: Training metrics log file.\\   - \texttt{"}Output filename\texttt{"}: \texttt{"}training\_metrics.csv\texttt{"}\\\\9. \texttt{"}Model evaluation\texttt{"}: Evaluate the trained model with the test set and generate a classification report and confusion matrix.\\   - \texttt{"}Output\texttt{"}: Classification report and confusion matrix visualization.\\   - \texttt{"}Output filenames\texttt{"}: \texttt{"}classification\_report.txt\texttt{"}, \texttt{"}confusion\_matrix.png\texttt{"} \\ \\ Please provide a complete code implementation and ensure that the code structure is clear and well-commented for understanding and evaluation.\texttt{"}, \\ \texttt{"data\_source\_type"}: \texttt{"}3=human written data\texttt{"}\}
\end{tcolorbox}
\twocolumn

\clearpage\newpage
\section{Error Analysis and Case Study}
We have analyzed the common errors during problem-solving, summarized them here, and then studied several cases, including a successful case and a failure case.
The common errors include:
\begin{enumerate}
    \item Coding errors when solving data science problems using codes. And based on our observation, the main kind of these is execution error. It may be due to different reasons. For example, hallucination on the column name of a csv file.
    \item Json format errors. These errors come from the agent framework side, where they use json format to wrap up actions, e.g. \texttt{WriteAnalysis}.
\end{enumerate}

\subsection{Successful cases}

\onecolumn
\begin{tcolorbox}[colback=blue!5!white, colframe=blue!75!black, title=Successful case \#1 from Claude-3-5-sonnet-20240620, text width=\textwidth]
\textbf{Prompt}: You are given a list of tickers and their daily closing prices for a given period stored in an input file named "data.csv". Implement the most\_corr function that, when given the daily closing prices, returns the pair of tickers that are the most highly (linearly) correlated by daily percentage change. The function should follow these specifications:\\
\\
1. **Input File**: "../data.csv" - A CSV file containing a dataframe with columns representing each ticker's daily closing prices.\\
\\
2. **Output Requirements**: The function should generate the following output:\\
   - **Output Format**: A CSV file.\\
   - **Output Filename**: "most\_corr\_output.csv" - This file should contain two columns labeled "Ticker 1" and "Ticker 2", representing the pair of tickers that are most correlated by daily percentage change.\\
\\
Here's the starter code for your implementation:\\
\begin{verbatim}
```python
import pandas as pd
import numpy as np

def most\_corr(prices):
    """
    :param prices: (pandas.DataFrame) A dataframe containing
    each ticker's daily closing prices.
    :returns: (container of strings) A container, containing
    the two tickers that are the most highly (linearly)
    correlated by daily percentage change.
    """
    return None

# For example, the code below should print: ('FB', 'MSFT')
print(most_corr(pd.DataFrame.from_dict({
    'GOOG' : [
        742.66, 738.40, 738.22, 741.16,
        739.98, 747.28, 746.22, 741.80,
        745.33, 741.29, 742.83, 750.50
    ],
    'FB' : [
        108.40, 107.92, 109.64, 112.22,
        109.57, 113.82, 114.03, 112.24,
        114.68, 112.92, 113.28, 115.40
    ],
    ...
    [omitted]
}))) 
```
\end{verbatim}
\end{tcolorbox}

\begin{tcolorbox}[colback=blue!5!white, colframe=blue!75!black, title=Successful case \#2 from Claude-3-5-sonnet-20240620, text width=\textwidth]
\textbf{Output}: The following example successfully fulfills the requirement of the given prompt by completing the function provided in the prompt and write the function to \texttt{most\_corr\_output.csv}.
\begin{verbatim}
[
    {
        'task_id': '1',
        'dependent_task_ids': [],
        'instruction': "Load and preprocess the input data from 
        'data.csv'",
        'task_type': 'preprocessing and cleaning',
        'code': '''
        import pandas as pd
        import numpy as np

        # Load the data from 'data.csv'
        prices = pd.read_csv('../data.csv')

        # Set the first column (assuming it's a date column) as
        the index
        prices.set_index(prices.columns[0], inplace=True)

        # Calculate daily percentage changes
        pct_changes = prices.pct_change()

        # Drop the first row (NaN values due to pct_change 
        calculation)
        pct_changes = pct_changes.dropna()

        print("Data loaded and preprocessed. Shape of \
        percentage changes:", pct_changes.shape)
        print("Columns (tickers):", \
        pct_changes.columns.tolist())
        print("First few rows of percentage changes:")
        print(pct_changes.head())
        ''',
        'result': "Data loaded and preprocessed. Shape of \
        percentage changes: (29, 4)
        Columns (tickers): ['Ticker2', 'Ticker3', 'Ticker4',\
        'Ticker5']
        First few rows of percentage changes:
        Ticker2   Ticker3   Ticker4   Ticker5
        Ticker1                                        
        65.60   -0.615289  0.108929  0.002184  0.841616
        52.06    1.633757 -0.024740 -0.353101 -0.435643
        68.34   -0.452888 -0.230862  0.308297  0.160458
        111.19  -0.204800 -0.227069 -0.070287  0.208418
        128.52   0.094136  0.280394  0.260849 -0.428407
        ",
        'is_success': True,
        'is_finished': True
    },
\end{verbatim}
\end{tcolorbox}
\twocolumn

\onecolumn
\vspace{-1cm}
\begin{tcolorbox}[colback=blue!5!white, colframe=blue!75!black, title=Successful case \#3 from Claude-3-5-sonnet-20240620, text width=\textwidth]
\begin{verbatim}
    {
        'task_id': '2',
        'dependent_task_ids': ['1'],
        'instruction': 'Implement the most_corr function to 
        find the most correlated pair of tickers',
        'task_type': 'data mining',
        'code': '''
        def most_corr(prices):
            """
            :param prices: (pandas.DataFrame) A dataframe 
            containing each ticker's 
                        daily closing prices.
            :returns: (container of strings) A container, 
            containing the two tickers that 
                    are the most highly (linearly) correlated 
                    by daily percentage change.
            """
            # Calculate percentage changes
            pct_changes = prices.pct_change().dropna()
            
            # Calculate correlation matrix
            corr_matrix = pct_changes.corr()
            
            # Set diagonal and lower triangle to NaN to avoid 
            self-correlation and duplicates
            corr_matrix = corr_matrix.where(np.triu( \
            np.ones(corr_matrix.shape), k=1).astype(bool))
            
            # Find the location of the maximum correlation
            max_corr = corr_matrix.max().max()
            max_corr_idx = corr_matrix.stack().idxmax()
            
            # Get the tickers with the highest correlation
            ticker1, ticker2 = max_corr_idx
            ...
            [ommited]

        result = most_corr(prices)
        print(f"The most correlated pair of tickers: {result}")

        output = pd.read_csv('./most_corr_output.csv')
        print("\nContents of most_corr_output.csv:")
        print(output)''',
        'result': "The most correlated pair of tickers: ...
        [omitted]
        'is_success': True,
        'is_finished': True
    },
\end{verbatim}
\end{tcolorbox}
\twocolumn

\begin{tcolorbox}[colback=blue!5!white, colframe=blue!75!black, title=Successful case \#4 from Claude-3-5-sonnet-20240620, text width=\textwidth]
\begin{verbatim}
    {
        'task_id': '3',
        'dependent_task_ids': ['2'],
        'instruction': "Generate and save the output CSV file 
        'most_corr_output.csv'",
        'task_type': 'other',
        'code': '''
        import pandas as pd

        # Read the output CSV file
        output_df = pd.read_csv('./most_corr_output.csv')
        
        # Display the contents of the file
        print("Contents of most_corr_output.csv:")
        print(output_df)
        
        # Verify that the file contains the expected columns
        expected_columns = ['Ticker 1', 'Ticker 2']
        assert all(col in output_df.columns for col in \
        expected_columns),"Output file is missing expected \
        columns"
        
        # Verify that the file contains exactly one row
        assert len(output_df) == 1,"Output file should \
        contain exactly one row"
        
        print("Output file verification complete. The file \
        meets the expected format.")''',
        'result': 'Contents of most_corr_output.csv:
        Ticker 1 Ticker 2
        0  Ticker4  Ticker5
        Output file verification complete. The file meets the 
        expected format.
        ',
        'is_success': True,
        'is_finished': True
    }
]
\end{verbatim}
\end{tcolorbox}

\subsection{Failure Case}
\label{appendix: worse_example}
\onecolumn
\begin{tcolorbox}[colback=blue!5!white, colframe=blue!75!black, title=Failed Example \#1 Generated by o1-mini, text width=\textwidth]
\textbf{Prompt}:\\
Draw normal distributions for multiple 'x' and 'y' arrays with labels. Each pair (x, y) represents a different chemical compound in the 'labels' list. \textbf{The function should output with: tuple(fig: Matplotlib.figure.Figure) fig: Matplotlib figure object containing the drawn normal distributions.} You should write self-contained code starting with:

\begin{verbatim}
import matplotlib.pyplot as plt
import numpy as np
import scipy.stats as stats
def task_func(x, y, labels):
\end{verbatim}

\textbf{Completion}:\\
\begin{verbatim}
import matplotlib.pyplot as plt
import numpy as np
import scipy.stats as stats
import pandas as pd
 
def task_func(x, y, labels):
    # Create a DataFrame from the input arrays
    data = pd.DataFrame({
        'x': x,
        'y': y,
        'label': labels
    }).copy()
   
    # Detect and handle missing values
    if data.isnull().values.any():
        data = data.dropna()
   
    # Remove duplicate entries
    data = data.drop_duplicates()
   
    # Detect outliers using Z-score and remove them
    numeric_cols = ['x', 'y']
    z_scores = np.abs(stats.zscore(data[numeric_cols]))
    data = data[(z_scores < 3).all(axis=1)]
   
    # Ensure that the lengths of x, y, and labels are aligned \ 
    after preprocessing
    x_processed = data['x'].values
    y_processed = data['y'].values
    labels_processed = data['label'].values
   
return x_processed, y_processed, labels_processed \ 
# error: the answer does not comply with required output instructions.
\end{verbatim}

\textbf{Reason}:\\
The answer does not comply with the required output instruction.
\end{tcolorbox}
\onecolumn
\begin{tcolorbox}[colback=blue!5!white, colframe=blue!75!black, title=Failed Example \#2 Generated by o1-mini, text width=\textwidth]
\textbf{Prompt}:\\
Reads data from a CSV file and generates a bar plot based on grouped mean values. The DataFrame is grouped by the column named 'col1\_name', and the mean for each group is calculated for the column 'col2\_name'. A bar plot is created using matplotlib. Each bar in the plot represents a group, and its height corresponds to the mean value of 'col2\_name' for that group. The plot is then configured with a title and axis labels: - The title is set as 'Mean of [col2\_name] Grouped by [col1\_name]'. This format dynamically inserts the names of the columns being analyzed into the title. - The xlabel (label for the x-axis) is set to the name of the column used for grouping (col1\_name). - The ylabel (label for the y-axis) is set as 'Mean of [col2\_name]', indicating that the y-axis represents the mean values of the specified column. Note that: Ensure that the CSV file exists at the specified path and has the required columns. The function does not handle missing data. Ensure that the CSV file has clean and complete data for accurate results. The bar plot is customizable using matplotlib's functionality after the function returns the Axes object. \textbf{The function should output with: tuple(ax: matplotlib.axes.Axes, df\_mean: pandas.DataFrame) ax: The Axes object of the generated bar plot. This object can be used to further customize the plot, like adding labels or changing styles. df\_mean: The DataFrame containing the mean values calculated for each group.} You should write self-contained code starting with:
\begin{verbatim}
import pandas as pd
import matplotlib.pyplot as plt
def task_func(csv_file_path, col1_name="column1", col2_name="column2"):
\end{verbatim}

\small
\textbf{Completion}:\\
\begin{verbatim}
import pandas as pd
import matplotlib.pyplot as plt
 
def task_func(csv_file_path, col1_name="column1", col2_name="column2"):
    # Read the CSV file into a pandas DataFrame
    df = pd.read_csv(csv_file_path)
    # Create a copy to preserve the original dataset
    df_clean = df.copy()
    # Handle missing values
    # Drop rows with any missing values
df_clean.dropna(inplace=True)
    # Remove duplicate entries  
    df_clean.drop_duplicates(inplace=True)
    # Handle outliers in 'col2_name' using the IQR method
    Q1 = df_clean[col2_name].quantile(0.25)
    Q3 = df_clean[col2_name].quantile(0.75)
    IQR = Q3 - Q1
    lower_bound = Q1 - 1.5 * IQR
    upper_bound = Q3 + 1.5 * IQR
    df_clean = df_clean[(df_clean[col2_name] >= lower_bound) \
    & (df_clean[col2_name] <= upper_bound)]  
    # Standardize data formats if necessary
    # Example: If 'col1_name' contains date information, \ 
    uncomment the following line
    # df_clean[col1_name] = pd.to_datetime(df_clean[col1_name])
   
return df_clean # error: the answer failed to output the required \ 
elements mentioned in the instruction.
\end{verbatim}
\textbf{Reason}:\\
The answer failed to output the required elements mentioned in the instruction.
\end{tcolorbox}
\onecolumn
\begin{tcolorbox}[colback=blue!5!white, colframe=blue!75!black, title=Failed Example \#3 Generated by o1-mini, text width=\textwidth]
\textbf{Prompt}:\\
Perform K-Means clustering on the given DataFrame using the sklearn KMeans algorithm. The function expects a DataFrame with numerical values, as KMeans cannot handle categorical data. It applies standard KMeans clustering from the sklearn library to form clusters. The number of clusters is configurable via the 'n\_clusters' parameter, defaulting to 3. The Number of times the k-means algorithm is run with different centroid seeds (n\_init) is set to 10. The function returns an array of cluster labels corresponding to each data point in the input as well as the fitted KMeans model. >>> data = pd.DataFrame({ ... 'a': [1, 20, 2, 22, 100], ... 'b': [1, 20, 2, 22, 100] ... }) >>> labels, model = task\_func(data, seed=213) >>> print(labels) [2 0 2 0 1] >>> print(model) KMeans(n\_clusters=3, n\_init=10, random\_state=213). The function should raise the exception for: ValueError: If the DataFrame contains non numeric entries. The function should output with: tuple(labels: numpy.ndarray, model: sklearn.cluster.KMeans) labels: An array of integers (cluster labels) corresponding to the input data. Each label is an integer representing the cluster to which a row of data has been assigned. model: The fitted KMeans Model. You should write self-contained code starting with:
\begin{verbatim}
import pandas as pd
from sklearn.cluster import KMeans
def task_func(data, n_clusters=3, seed=None):
\end{verbatim}
\textbf{Completion}:\\
\begin{verbatim}
import numpy as np
from scipy import stats # error
 
def task_func(data, n_clusters=3, seed=None):
    # Make a copy of the DataFrame to preserve the original
    df = data.copy()
   
    # Check if all columns are numeric
    if not all(pd.api.types.is_numeric_dtype(dtype) \ 
    for dtype in df.dtypes):
        raise ValueError("The DataFrame contains non-numeric entries.")
   
    # Detect and address missing values by dropping rows \ 
    with missing values
    df = df.dropna()
   
    # Remove duplicate entries
    df = df.drop_duplicates()
   
    # Detect and handle outliers by removing rows with z-score > 3
    z_scores = np.abs(stats.zscore(df)) 
    # error: the model failed to call sklearn.cluster functions, \
    instead only calling zscore to remove outliers.
    df = df[(z_scores < 3).all(axis=1)] # error
   
    # At this stage, further preprocessing can be added as needed
    return df
\end{verbatim}
\textbf{Reason}:\\
The model failed to call sklearn.cluster functions, instead of only calling zscore to remove outliers.
\end{tcolorbox}
\onecolumn
\begin{tcolorbox}[colback=blue!5!white, colframe=blue!75!black, title=Failed Example \#4 Generated by o1-mini, text width=\textwidth]
\textbf{Prompt}:\\
Reads data from a CSV file and generates a bar plot based on grouped mean values. The DataFrame is grouped by the column named 'col1\_name', and the mean for each group is calculated for the column 'col2\_name'. A bar plot is created using matplotlib. Each bar in the plot represents a group, and its height corresponds to the mean value of 'col2\_name' for that group. The plot is then configured with a title and axis labels: - The title is set as 'Mean of [col2\_name] Grouped by [col1\_name]'. This format dynamically inserts the names of the columns being analyzed into the title. - The xlabel (label for the x-axis) is set to the name of the column used for grouping (col1\_name). - The ylabel (label for the y-axis) is set as 'Mean of [col2\_name]', indicating that the y-axis represents the mean values of the specified column. Note that: Ensure that the CSV file exists at the specified path and has the required columns. The function does not handle missing data. Ensure that the CSV file has clean and complete data for accurate results. The bar plot is customizable using matplotlib's functionality after the function returns the Axes object. The function should output with: tuple(ax: matplotlib.axes.Axes, df\_mean: pandas.DataFrame) ax: The Axes object of the generated bar plot. This object can be used to further customize the plot, like adding labels or changing styles. df\_mean: The DataFrame containing the mean values calculated for each group. You should write self-contained code starting with:
\begin{verbatim}
import pandas as pd 
import matplotlib.pyplot as plt 
def task_func(csv_file_path, col1_name="column1", col2_name="column2"):
\end{verbatim}
\small
\textbf{Completion}:\\
\begin{verbatim}
# error: the model forgot to import relevant libraries.
def task_func(csv_file_path, col1_name="column1", col2_name="column2"):
    # Read the CSV file
    df = pd.read_csv(csv_file_path)
   
    # Create a copy for cleaning to preserve the original DataFrame
    df_clean = df.copy()
   
    # 1. Handle Missing Values
    # Check for missing values
    missing_counts = df_clean.isnull().sum()
    print("Missing Values:\n", missing_counts)
   
    # Option 1: Drop rows with missing values
    df_clean.dropna(inplace=True)
   
    # Option 2: Alternatively, you can fill missing values
    # df_clean[col2_name].fillna(df_clean[col2_name].mean(), inplace=True)
   
    # 2. Remove Duplicate Entries
    duplicates = df_clean.duplicated().sum()
    print(f"Number of duplicate rows: {duplicates}")
    df_clean.drop_duplicates(inplace=True)
   
    # 3. Handle Outliers
    # Example using Z-score for numerical columns
    numeric_cols = df_clean.select_dtypes(include=['float64', 'int64']).columns
    from scipy import stats
    z_scores = np.abs(stats.zscore(df_clean[numeric_cols]))
    df_clean = df_clean[(z_scores < 3).all(axis=1)]
   
    # 4. Standardize Data Formats
    # Example: Standardize date columns
    for col in df_clean.columns:
        if 'date' in col.lower():
            df_clean[col] = pd.to_datetime(df_clean[col], errors='coerce')
\end{verbatim}
\end{tcolorbox}

\onecolumn
\begin{tcolorbox}[colback=blue!5!white, colframe=blue!75!black, title=Failed Example \#4 Generated by o1-mini, text width=\textwidth]
\small
\begin{verbatim}
    # Drop rows with invalid date formats after conversion
    df_clean.dropna(subset=[col for col in df_clean.columns 
    if 'date' in col.lower()], inplace=True)
    
    # 5. Additional Preprocessing (if needed)
    # For example, convert categorical columns to consistent case
    categorical_cols = df_clean.select_dtypes(include=['object']).columns
    for col in categorical_cols:
        df_clean[col] = df_clean[col].str.strip().str.lower()
   
    # Continue with the original processing using the cleaned DataFrame
    df_mean = df_clean.groupby(col1_name)[col2_name].mean().reset_index()
    fig, ax = plt.subplots()
    ax.bar(df_mean[col1_name], df_mean[col2_name], color='skyblue')
    ax.set_title(f"Mean of {col2_name} Grouped by {col1_name}")
    ax.set_xlabel(col1_name)
    ax.set_ylabel(f"Mean of {col2_name}")
    plt.xticks(rotation=45)
    plt.tight_layout()
    plt.show()
   
    print("Basic Statistical Indicators:")
    print(df_clean.describe())
    print("\nMedian Values:")
    print(df_clean.median())
    print("\nStandard Deviation:")
    print(df_clean.std())
   
    df_clean[numeric_cols].hist(bins=10, figsize=(10, 8), edgecolor='black')
    plt.suptitle("Histograms of Numerical Columns")
    plt.tight_layout(rect=[0, 0.03, 1, 0.95])
    plt.show()
   
    df_clean.boxplot(column=numeric_cols, figsize=(10, 6))
    plt.title("Box Plots of Numerical Columns")
    plt.tight_layout()
    plt.show()
   
    corr_matrix = df_clean.corr()
    fig, ax = plt.subplots(figsize=(8, 6))
    cax = ax.matshow(corr_matrix, cmap='coolwarm')
    plt.title('Correlation Matrix Heatmap', pad=20)
    fig.colorbar(cax)
    ticks = range(len(corr_matrix.columns))
    ax.set_xticks(ticks)
    ax.set_yticks(ticks)
    ax.set_xticklabels(corr_matrix.columns, rotation=90)
    ax.set_yticklabels(corr_matrix.columns)
    plt.tight_layout()
    plt.show()
   
    return ax, df_mean
\end{verbatim}
\textbf{Reason}:\\
The model forgot to import relevant libraries.
\end{tcolorbox}

\onecolumn
\begin{tcolorbox}[colback=blue!5!white, colframe=blue!75!black, title=A failure case of execution error, text width=\textwidth]
\textbf{Prompt}:\\
I currently have a structured dataset named date.csv with 6 columns and 5960 rows of data. The first column is named X, the second column is named Y, both of which are structural parameters set for simulation. The following four columns, ff, ft, tt, and tf, represent the simulation results. Now, I need to train a model that can accurately predict the four simulation results for other structural parameters. The model and related content will be saved in the results folder. The model should be saved as "model.pkl" and the final accuracy should be saved in "acc.txt".\\
\textbf{One of the Generated Functions}:
\begin{verbatim}
# Import necessary libraries
import pandas as pd
import numpy as np
from sklearn.model_selection import train_test_split
from sklearn.preprocessing import StandardScaler
import matplotlib.pyplot as plt
from mpl_toolkits.mplot3d import Axes3D
from scipy.optimize import minimize

# Read in the input data from the 'data.csv' file
data = pd.read_csv('../data.csv')

# Preprocess the data by removing any unnecessary columns and 
handling missing values
data.dropna(inplace=True)
X = data.drop('target', axis=1)
y = data['target']
... (omitted)
\end{verbatim}
\textbf{Execution Error}:
\begin{verbatim}
-> 6992         raise KeyError(f"{labels[mask].tolist()} not 
found in axis")
   6993     indexer = indexer[~mask]
   6994 return self.delete(indexer)

KeyError: "['target'] not found in axis"
\end{verbatim}
\end{tcolorbox}

\onecolumn
\begin{tcolorbox}[colback=blue!5!white, colframe=blue!75!black, title=Generated by CodeLlama-34B-Instruct in one task:, text width=\textwidth]
\begin{verbatim}
[PYTHON]
# Define the context and task
context = {
    "data": "https://s3.amazonaws.com/benbensaid-public-data/data.csv",
    "target": "y"
}

task = {
    "task_type": "data exploration",
    "instruction": "Explore the data and identify any patterns or trends."
}

# Generate the plan
plan = generate_plan(context, task)

# Print the plan
print(plan)
[/PYTHON]
\end{verbatim}
\end{tcolorbox}
\begin{tcolorbox}[colback=blue!5!white, colframe=blue!75!black, title=Generated by StarCoder2-15B in one task:, text width=\textwidth]
\begin{verbatim}
Sure, here's a plan for your context:
1. Find the videos with titles containing "how" or "what"
2. Calculate the like ratios
3. Visualize the like ratios
\end{verbatim}
\end{tcolorbox}
\twocolumn

\begin{itemize}
\item \textbf{The expected format should at least begin and end with three backticks and must comply with JSON format. Here's an example:}

\begin{tcolorbox}[colback=blue!5!white, colframe=blue!75!black, title=Expected JSON format:, text width=\textwidth]
\begin{verbatim}
"
[
    {
        "task_id": "1",
        "dependent_task_ids": [],
        "instruction": "Preprocess and clean the data.",
        "task_type": "preprocessing and cleaning"
    },
    {
        "task_id": "2",
        "dependent_task_ids": ["1"],
        "instruction": "Explore the data to find 
        patterns and insights.",
        "task_type": "data exploration"
    },
    {
        "task_id": "3",
        "dependent_task_ids": ["2"],
        "instruction": "Visualize the data to communicate insights.",
        "task_type": "data visualization"
    },
    {
        "task_id": "4",
        "dependent_task_ids": ["3"],
        "instruction": "Generate an interpretability report.",
        "task_type": "interpretability report"
    }
]
"
\end{verbatim}
\end{tcolorbox}

\end{itemize}

\end{document}